\documentclass[smallextended]{svjour3}       
\smartqed  
\usepackage{natbib}

\usepackage{graphicx}
\usepackage{url}

\usepackage[ruled, vlined, linesnumbered]{algorithm2e}
\DontPrintSemicolon                        
\SetKwInput{KwInput}{Input}                
\SetKwInput{KwOutput}{Output}              
\let\oldnl\nl
\newcommand{\nonl}{\renewcommand{\nl}{\let\nl\oldnl}}

\usepackage{textcomp}
\usepackage{xcolor}
\usepackage{csquotes}
\usepackage{booktabs}

\usepackage{amsmath,amsfonts}
\usepackage{bm}
\usepackage{dsfont}
\usepackage{float} 

\newcommand{\Xspace}{\mathcal{X}}                                           
\newcommand{\D}{\mathcal{D}}                                                
\renewcommand{\xi}[1][i]{x^{(#1)}}                                          
\newcommand{\yi}[1][i]{y^{(#1)}}                                            
\newcommand{\xyi}{\left(\xi, \yi\right)}                                    
\newcommand{\xjb}{\mathbf{x}_j}                                             

\newcommand{\fh}{\hat{f}}                                                   
\ifdefined\N                                                                
\renewcommand{\N}{\mathds{N}}                                                
\else
  \newcommand{\N}{\mathds{N}}
\fi
\newcommand{\R}{\mathds{R}}                                                 
\ifdefined\C
  \renewcommand{\C}{\mathds{C}}                                             
\else
  \newcommand{\C}{\mathds{C}}
\fi

\newcommand{\xv}{\bm{x}}													

\renewcommand{\P}{\mathds{P}}                                               
\newcommand{\E}{\mathds{E}}                                                 

\newcommand{\xij}{x_{j}^{(i)}}        
\newcommand{\indep}{\rotatebox[origin=c]{90}{$\models$}}

\newcommand{\orcidID}[1]{}

\begin{document}

\title{Model-agnostic Feature Importance and Effects with Dependent Features--A Conditional Subgroup Approach}
\titlerunning{Importance and Effects with Dependent Features}
\author{Christoph Molnar \orcidID{0000-0003-2331-868X} \and
        Gunnar K\"{o}nig \orcidID{0000-0001-6141-4942} \and
        Bernd Bischl \orcidID{0000-0001-6002-6980} \and
        Giuseppe Casalicchio\orcidID{0000-0001-5324-5966} }
\institute{
       C. Molnar$^1$, G. K\"{o}nig$^2$, B. Bischl$^3$, G. Casalicchio$^4$ \at 
       Department of Statistics, Ludwig-Maximilians-University Munich, Munich, Germany
       \and
       C. Molnar$^1$ \at
       Leibniz Institute for Prevention Research and Epidemiology – BIPS GmbH, Bremen, Germany \\
       \email{christoph.molnar.ai@gmail.com}
       \and
       {CRediT taxonomy: 
       Conceptualization: 1, 2, 3, 4; Methodology: 1, 2, 4; Formal analysis and investigation: 1, 2; Writing - original draft preparation: 1, 2; Writing - review and editing: 2, 3, 4;   Visualization: 1; Validation: 1, 2; Software: 1; Funding acquisition: 1, 3, 4; Supervision: 3, 4}
       }

\date{}

\maketitle


\begin{abstract}%
The interpretation of feature importance in machine learning models is challenging when features are dependent.
Permutation feature importance (PFI) ignores such dependencies, which can cause misleading interpretations due to extrapolation.
A possible remedy is more advanced conditional PFI approaches that enable the assessment of feature importance conditional on all other features. 
Due to this shift in perspective and in order to enable correct interpretations, it is therefore important that the conditioning is transparent and humanly comprehensible.
In this paper, we propose a new sampling mechanism for the conditional distribution based on permutations in conditional subgroups.
As these subgroups are constructed using decision trees (transformation trees), the conditioning becomes inherently interpretable.
This not only provides a simple and effective estimator of conditional PFI, but also local PFI estimates within the subgroups.
In addition, we apply the conditional subgroups approach to partial dependence plots (PDP), a popular method for describing feature effects that can also suffer from extrapolation when features are dependent and interactions are present in the model.
We show that PFI and PDP based on conditional subgroups often outperform methods such as conditional PFI based on knockoffs, or accumulated local effect plots.
Furthermore, our approach allows for a more fine-grained interpretation of feature effects and importance within the conditional subgroups.
\end{abstract}


\section{Introduction}
\label{sec:introduction}

\begin{figure}
  \includegraphics[width=\columnwidth]{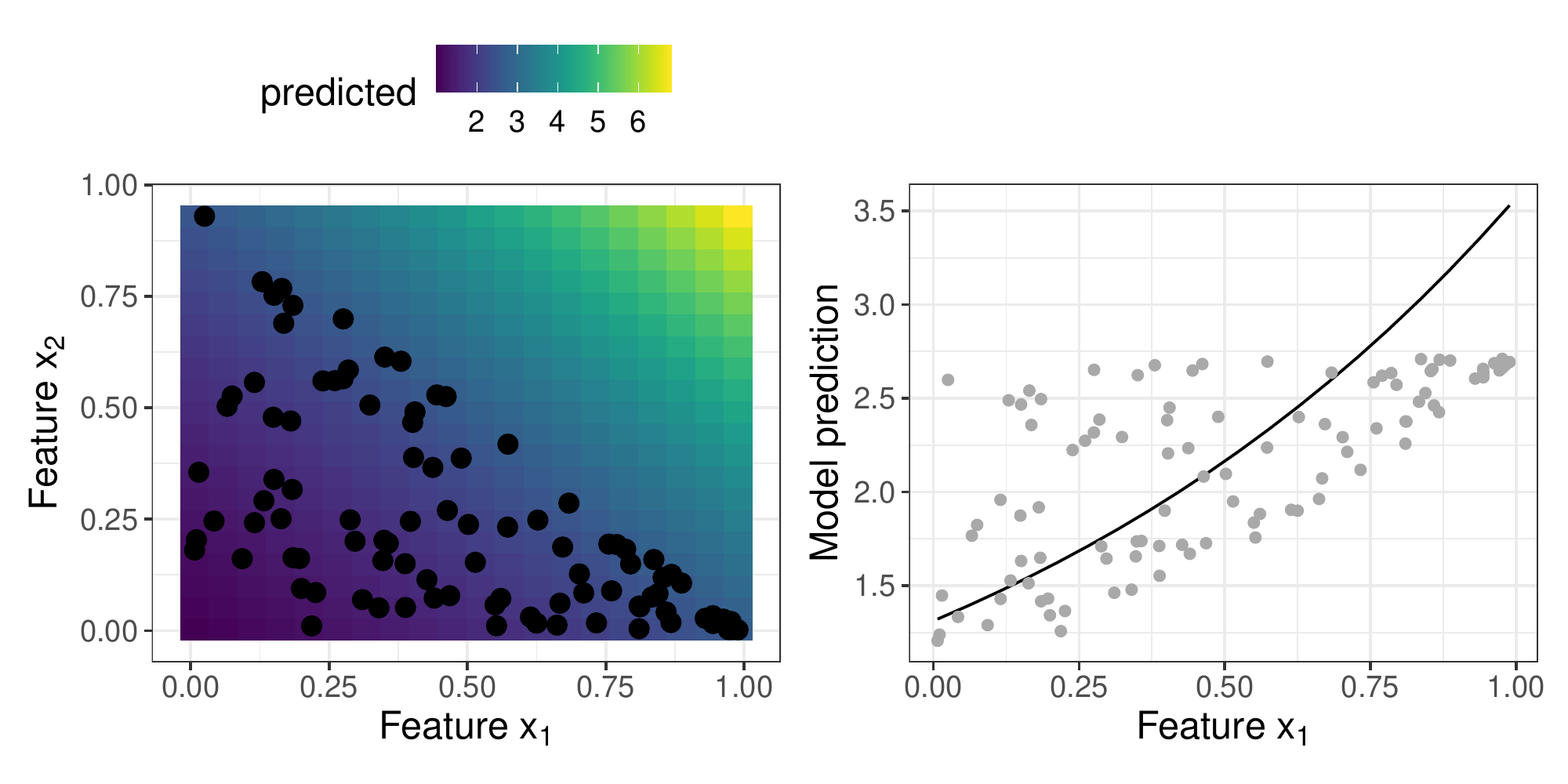} \caption{Misleading PDP. Simulation of features $x_1 \sim U(0,1)$, $x_2 \sim U(0, 1-x_1)$ and a non-additive prediction model $\fh(x) = exp(x_1 + x_2)$. \textbf{Left:} Scatter plot with 100 data points and the prediction surface of $\fh$. \textbf{Right:} PDP of $x_1$. The grey dots are observed $(x_1,\fh(x_1,x_2))$-pairs. For $x_1 > 0.75$ the PDP suggests higher average predictions than the maximum prediction observed in the data.}\label{fig:correlation-problem}
\end{figure}

Many model-agnostic machine learning (ML) interpretation methods (see \cite{molnar2019,guidotti2018survey} for an overview) are based on making predictions on perturbed input features, such as permutations of features.
The partial dependence plot (PDP) \citep{friedman1991multivariate} visualizes how changing a feature affects the prediction on average.
The permutation feature importance (PFI) \citep{breiman2001random,fisher2019all} quantifies the importance of a feature as the reduction in model performance after permuting a feature.
PDP and PFI change feature values without conditioning on the remaining features.
If features are dependent, such changes can lead to extrapolation to areas of the feature space with low density.
For non-additive models such as tree-based methods or neural networks, extrapolation can result in misleading interpretations \citep{strobl2008conditional,Tolosi2011,hooker2019please,molnar2020pitfalls}.
An illustration of the problem is given in Figure~\ref{fig:correlation-problem}.
%

Extrapolation can be avoided by sampling a feature conditional on all other features and thereby preserving the joint distribution \citep{strobl2008conditional,hooker2019please}.
This yields conditional variants of the PDP and PFI that have to be interpreted differently.
While the interpretation of marginal PDP and PFI is \textit{independent} of the other features, the interpretation of conditional PDP and PFI is \textit{conditional} on other features.

\begin{figure}
\centering
  \includegraphics[width=0.8\textwidth]{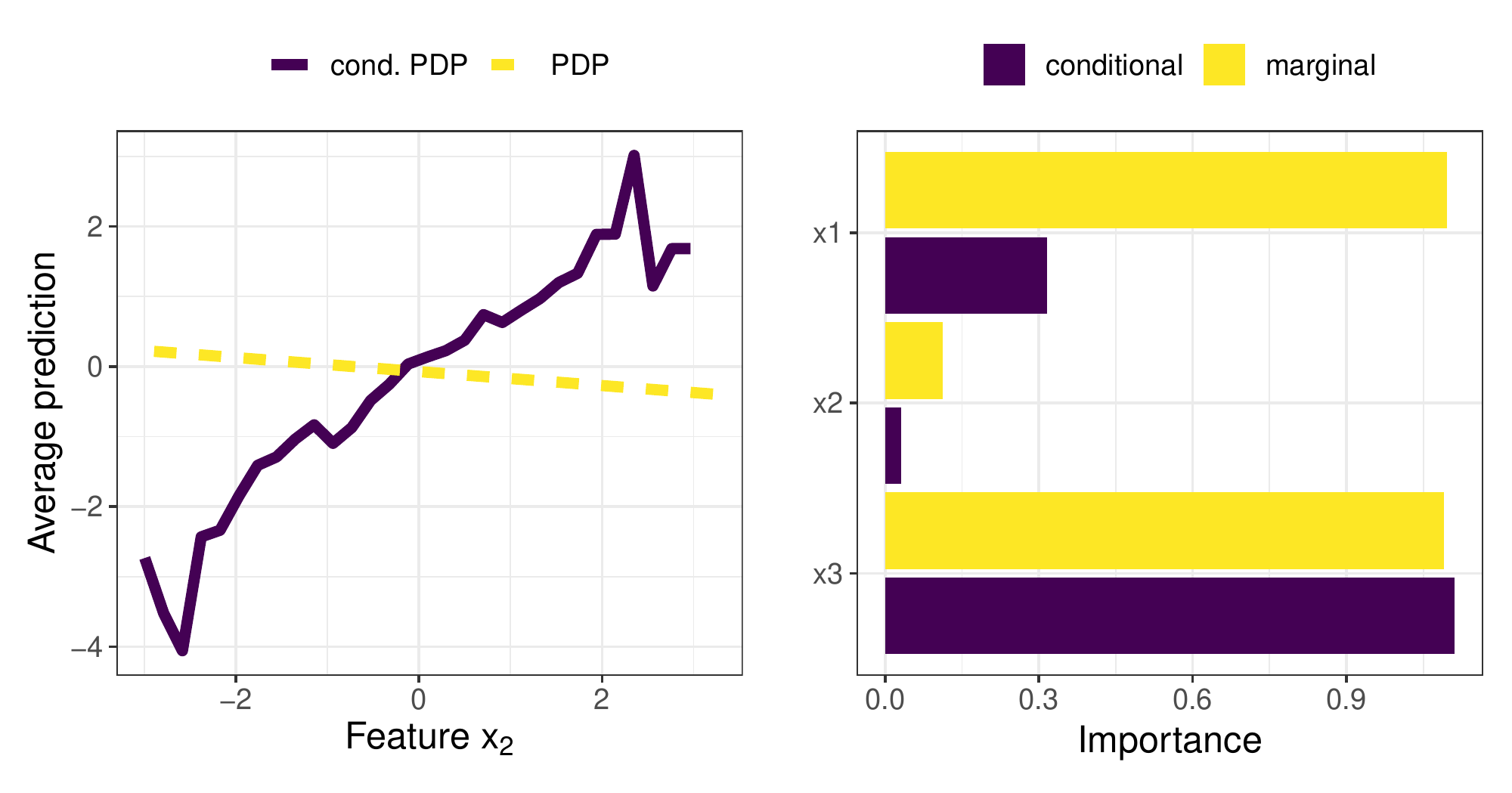} \caption{Simulation of a linear model $\fh(x) = x_1 - 0.1 \cdot x_2 + x_3$ with $x_1,x_2, x_3 \sim N(0,1)$ and a correlation of 0.978 between $x_1$ and $x_2$. \textbf{Left:} PDP and conditional PDP for feature $x_2$. The conditional PDP mixes the effects of $x_1$ and $x_2$ and thus shows a positive effect.
  \textbf{Right:} PFI and conditional PFI of $x_1$, $x_2$ and $x_3$. The PFI of $x_1$ decreases when $x_1$ is permuted conditional on $x_2$ and vice versa.}\label{fig:mplot-fail}
\end{figure}

Figure~\ref{fig:mplot-fail} shows how conditional PFI can be misinterpreted:
Features $X_1$ and $X_3$ have the same coefficient in a linear model and  the same marginal PFI, but $X_1$ has a lower conditional PFI since it is correlated with feature $X_2$.
The conditional PFI must be interpreted as the additional, unique contribution of a feature given all features we conditioned on \citep{konig2020relative,fisher2019all}.
It therefore has also been called \enquote{partial importance} \citep{debeer2020conditional}.
If interpreted incorrectly, this can lead to the wrong conclusion that, for example, two strongly dependent features are irrelevant for the prediction model (Figure~\ref{fig:mplot-fail}).
The correct conclusion would be that a feature is less relevant given knowledge of the dependent feature.

In Figure~\ref{fig:mplot-fail}, the conditional PDP shows a positive effect for a feature that has a negative coefficient in a linear regression model.
The discrepancy is due to correlation of the feature with another feature with a large positive coefficient.
The conditional effect of a feature is a mix of its marginal effect and the marginal effects of all dependent features \citep{hooker2019please,apley2016visualizing}.
While conditional PFI might assign a low importance to a feature on which the model relied heavily, the conditional PDP has the opposite pitfall: it can show an effect for a feature that was not used by the model.
This interpretation might be undesirable and is similar to the omitted variable bias phenomenon, which also happens in Figure~\ref{fig:mplot-fail}: regressing $\fh$ from $X_2$, while ignoring $X_1$ \citep{apley2016visualizing}.

The interpretation of conditional PFI and PDP requires knowledge of the dependence structure between the feature of interest and the other features.
Such knowledge of dependence structures would help explain differences between a feature's marginal and conditional PFI and break down the conditional PDP into the effect of the feature of interest and that of the dependent features.
However, state-of-the-art conditional sampling mechanisms such as knockoffs \citep{barber2015controlling,candes2018panning,watson2019testing} do not provide a readily interpretable conditioning.

%
Our \textbf{contributions} are the following.
We propose the \textbf{c}onditional \textbf{s}ubgroup PDPs (cs-PDPs) and PFIs (cs-PFIs).
Both are based on conditional subgroup permutation (cs-permutation), a sampling method for the conditional distribution.
Standard (i.e., marginal) PDPs and PFIs are computed and interpreted within subgroups of the data, enabling a local interpretation of feature effect and importance while handling the problem of extrapolation.
We construct the subgroups for a feature by training a decision tree in which the distribution of the feature becomes less dependent on other features.
The tree structure allows interpretation of how other features influence the effect and importance of the feature at hand.
We show that the conditional PFI estimate based on cs-PFIs can recover the ground truth in simulations and often outperforms related approaches.
In addition, we study how well different conditional PDP/PFI approaches retain the joint distribution of data sets from the OpenML-CC18 benchmarking suite \citep{bischl2019openml} and show that cs-permutation achieves state-of-the-art data fidelity.
We demonstrate that the cs-PDPs have a high model fidelity, that is, they are closer to the model prediction than other feature effect methods.
By inspecting the cs-PFIs and cs-PDPs in combination with the respective subgroup descriptions, insights into the model and the dependence structure of the data are possible.
We show how we can trade off human-intelligibility of the subgroups for extrapolation by choosing the granularity of the grouping.
In an application, we illustrate how cs-PDPs and cs-PFIs can reveal new insights into the ML model and the data.
\section{Notation and Background}
\label{sec:notation}
We consider ML prediction functions $\fh:\R^p \mapsto \R$, where $\fh(\xv)$ is a model prediction and $\xv \in \R^p$ is a $p$-dimensional feature vector.
We use $\xjb \in \R^n$ to refer to an observed feature (vector) and $X_j$ to refer to the $j$-th feature as a random variable.
With $\mathbf{x}_{-j}$ we refer to the complementary feature space $\mathbf{x}_{\{1, \ldots, p\} \setminus \{j\}} \in \R^{n\times (p-1)}$ and  with $X_{-j}$ to the corresponding random variables.
We refer to the value of the $j$-th feature from the $i$-th instance as $\xij$ and to the tuples $\D = \{\xyi\}_{i=1}^n$ as data.

The \textbf{Permutation Feature Importance (PFI)} is defined as the increase in loss when feature $X_j$ is permuted:
\begin{equation}
 PFI_j = \E [L(Y, \fh(\tilde{X}_j, X_{-j}))] - \E[L(Y, \fh(X_j, X_{-j}))]
 \end{equation}

If the random variable $\tilde{X}_j$ has the same marginal distribution as $X_j$ (e.g., permutation), the estimate yields the marginal PFI.
If $\tilde{X}_j$ follows the conditional distribution $\tilde{X}_j \sim X_j | X_{-j}$, we speak of the conditional PFI.
The PFI is estimated with the following formula:
\begin{equation}
  \widehat{PFI}_j = \frac{1}{n} \sum_{i=1}^n\left(\frac{1}{M}\sum_{m=1}^M \tilde{L}^{m(i)} - L^{(i)})\right)
\label{eq:pfi}
\end{equation}
where $L^{(i)}= L(\yi, \fh(\xv^{(i)}))$ is the loss for the $i$-th observation and  $\tilde{L}^{(i)}=L(y^{(i)}, \fh(\tilde{x}_{j}^{(i)},\xv_{-j}^{(i)}))$ is the loss where $x_{j}^{(i)}$ was replaced by the m-th sample $\tilde{x}_{j}^{m(i)}$.
The latter refers to the $i$-th feature value obtained by a sample of $\mathbf{x}_j$. 
The sample can be repeated $M$-times for a more stable estimation of $\tilde{L}^{(i)}$.
Numerous variations of this formulation exist.
\cite{breiman2001random} proposed the PFI for random forests, which is computed from the out-of-bag samples of individual trees.
Subsequently, \cite{fisher2019all} introduced a model-agnostic PFI version.

The marginal \textbf{Partial Dependence Plot (PDP)} \citep{friedman1991multivariate} describes the average effect of the j-th feature on the prediction.
\begin{align}
PDP_j(x) = \E [\fh(x, X_{-j})],
\label{eq:theoretical-pdp}
\end{align}
If the expectation is conditional on $X_j$, $\E[\fh(x, X_{-j})| X_j = x]$, we speak of the conditional PDP.
The marginal PDP evaluated at feature value $x$ is estimated using Monte Carlo integration:
\begin{equation}
  \widehat{PDP}_{j}(x)=\frac{1}{n}\sum_{i=1}^n \fh(x,\xv^{(i)}_{-j})
  \label{eq:pdp}
\end{equation}
\section{Related Work}
\label{sec:related}
In this section, we review conditional variants of PDP and PFI and other approaches that try to avoid extrapolation.
%
\subsection{Related Work on Conditional PDP}

The marginal plot (M-Plot) \citep{apley2016visualizing} averages the predictions locally on the feature grid and mixes effects of dependent features.

\cite{hooker2007generalized} proposed a functional ANOVA decomposition with hierarchically orthogonal components, based on integration using the joint distribution of the data, which in practice is difficult to estimate.

Accumulated Local Effect (ALE) plots by \cite{apley2016visualizing} reduce extrapolation by accumulating the finite differences computed within intervals of the feature of interest.
By definition, interpretations of ALE plots are thus only valid locally within the intervals.
Furthermore, there is no straightforward approach to derive ALE plots for categorical features, since ALE requires ordered feature values.
Our proposed approach can handle categorical features.

Another PDP variant based on stratification was proposed by \cite{parr2019stratification}.
However, this stratified PDP describes only the data and is independent of the model.

Individual Conditional Expectation (ICE) curves by \cite{goldstein2015peeking} can be used to visualize the interactions underlying a PDP, but they also suffer from the extrapolation problem.
The \enquote{conditional} in ICE refers to conditioning on individual observations and not on certain features.
As a solution, \cite{hooker2019please} suggested to visually highlight the areas of the ICE curves in which the feature combinations are more likely.

\subsection{Related Work on Conditional PFI}

We review approaches that modify the PFI \citep{breiman2001random,fisher2019all} in presence of dependent features by using a conditional sampling strategy.

\cite{strobl2008conditional} proposed the conditional variable importance for random forests (CVIRF), which is a conditional PFI variant of \cite{breiman2001random}.
CVIRF was further analyzed and extended by \cite{debeer2020conditional}.
Both CVIRF and our approach rely on permutations based on partitions of decision trees.
However, there are fundamental differences.
CVIRF is specifically developed for random forests and relies on the splits of the underlying individual trees of the random forest for the conditional sampling.
In contrast, our cs-PFI approach trains decision trees for each feature using $X_{-j}$ as features and $X_j$ as the target.
Therefore, the subgroups for each feature are constructed from their conditional distributions (conditional on the other features) in a separate step, which is decoupled from the machine learning model to be interpreted.
Our cs-PFI approach is model-agnostic, independent of the target to predict and not specific to random forests.

\cite{hooker2019please} made a general suggestion to replace feature values by estimates of $\E[X_j|X_{-j}]$.

\cite{fisher2019all} suggested to use matching and imputation techniques to generate samples from the conditional distribution.
If $X_{-j}$ has few unique combinations, they suggested to group $\xij$ by unique $\xv_{-j}^{(i)}$ combinations and permute them for these fixed groups.
For discrete and low-dimensional feature spaces, they suggest non-parametric matching and weighting methods to replace $X_j$ values.
For continuous or high-dimensional data, they suggest imputing $X_j$ with $\E[X_j|X_{-j}]$ and adding residuals (under the assumption of homogeneous residuals).
Our approach using permutation in subgroups can be seen as a model-driven, binary weighting approach extended to continuous features.

Knockoffs \citep{candes2018panning} are random variables which are \enquote{copies} of the original features that preserve the joint distribution but are otherwise independent of the prediction target.
Knockoffs can be used to replace feature values for conditional feature importance computation.
\cite{watson2019testing} developed a testing framework for PFI based on knockoff samplers such as Model-X knockoffs \citep{candes2018panning}.
Our approach is complementary since \cite{watson2019testing} is agnostic to the sampling strategy that is used.
Others have proposed to use generative adversarial networks for generating knockoffs \citep{romano2019deep}.
Knockoffs are not transparent with respect to how they condition on the features, while our approach creates interpretable subgroups.

Conditional importance approaches based on model retraining have been proposed \citep{hooker2019please,lei2018distribution,gregorutti2017correlation}.
Retraining the model can be expensive, and answers a fundamentally different question, often related to feature selection and not based on a fixed set of features.
Hence, we focus on approaches that compute conditional PFI for a fixed model without retraining.

None of the existing approaches makes the dependence structures between the features explicit.
It is unclear which of the features in $X_{-j}$ influenced the replacement of $X_j$ the most and how.
Furthermore, little attention has been paid on evaluating how well different sampling strategies address the extrapolation problem.
We address this gap with an extensive data fidelity experiment on the OpenML-CC18 benchmarking suite.
To the best of our knowledge, our paper is also the first to conduct experiments using ground truth for the conditional PFI.
Our approach works with any type of feature, be it categorical, numerical, ordinal and so on, since we rely on decision trees to find the subgroups used for conditioning.
The differences between the different (conditional) PDP and PFI approaches ultimately boil down to how they sample from the conditional distribution.
Table \ref{tab:perturbations} lists different sampling strategies of model-agnostic interpretation methods and summarizes their assumptions to preserve the joint distribution.
\begin{table}
  \begin{tabular}{|p{0.3\textwidth}|p{0.4\textwidth}|p{0.2\textwidth}|}
    \hline
    Sampling Strategy & Used/Suggested By & Assumptions  \\
    \hline
    \hline
    No intervention on $X_j$ & Drop-and-Refit, \linebreak LOCO \citep{lei2018distribution} &  \\
    \hline
    Permute $X_j$ & Marginal PFI \citep{breiman2001random,fisher2019all}, \linebreak PDP \citep{friedman1991multivariate}   & $X_j \indep X_{-j}$ \\
    \hline
    Replace $X_j$ by knockoff $Z_j$ with $(Z_j, X_{-j})\sim (X_j,X_{-j})$ and $Z_j \indep Y$ & Knockoffs \citep{candes2018panning}, \linebreak CPI \citep{watson2019testing} & $(X_j, X_{-j}) \sim N$ \\
    \hline
    Move each $\xij$ to left and right interval bounds  & ALE \citep{apley2016visualizing} & $X_j \indep X_{-j}$ in intervals  \\
    \hline
    Permute $X_j$ in subgroups & cs-PFI, cs-PDP & $X_j \indep X_{-j}$ in subgroups   \\
    \hline
    Permute $X_j$ in random forest tree nodes & CVIRF \citep{strobl2008conditional,debeer2020conditional} &  $X_j \indep X_{-j}$ cond. on tree splits in $X_{-j}$ to predict $Y$ \\
    \hline
    Impute $X_j$ from $X_{-j}$ & \citep{fisher2019all} & Homogeneous residuals  \\
    \hline
\end{tabular}
  \caption{Sampling strategies for model-agnostic interpretation techniques.}
  \label{tab:perturbations}
\end{table}

\section{Conditional Subgroups}
\label{sec:conditional-models}
\begin{figure}
\centering
\includegraphics[width=0.9\columnwidth, clip = TRUE, trim = 0cm 0.5cm 0 0.2cm ]{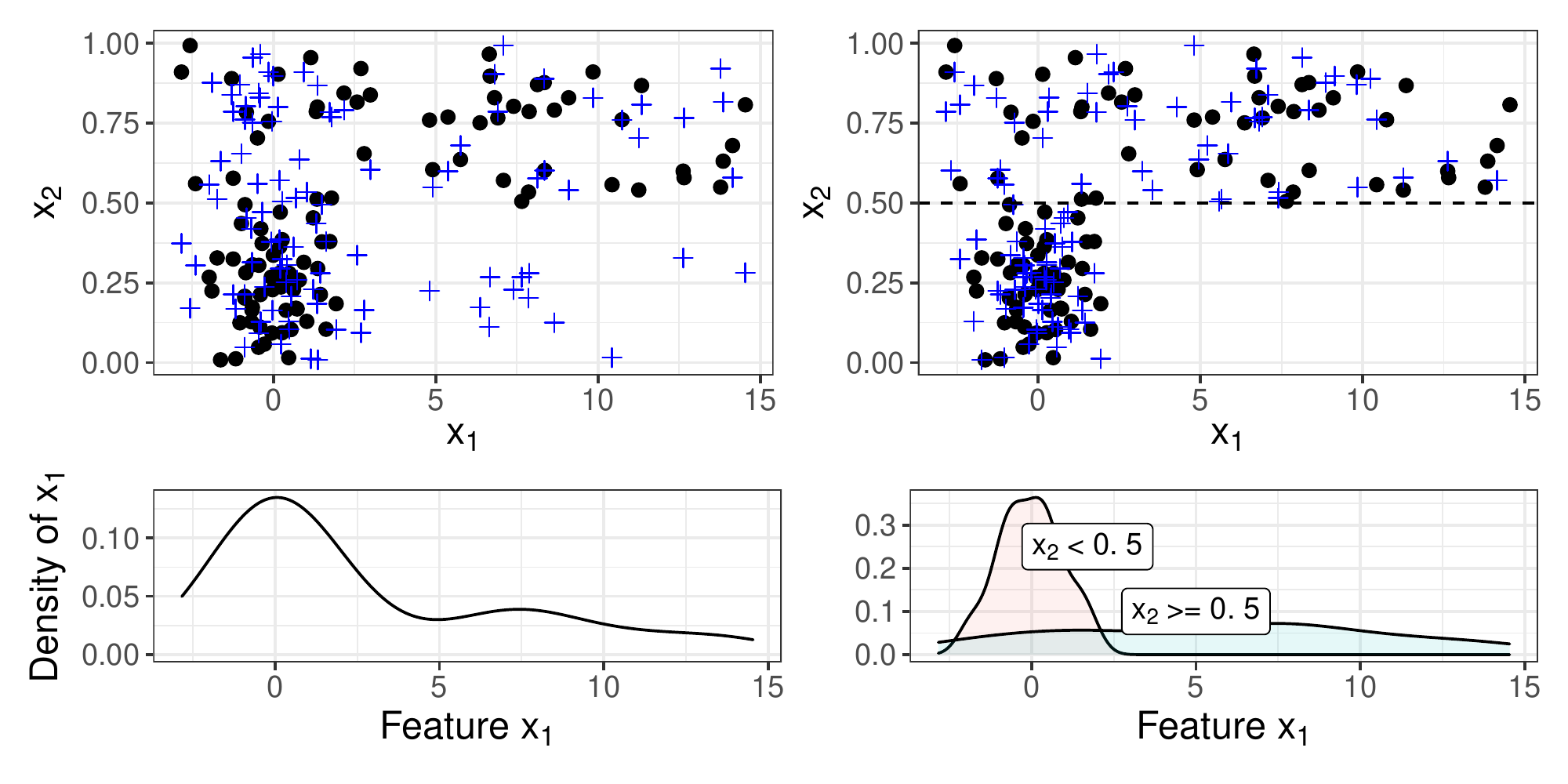} \caption{Features $X_2 \sim U(0,1)$ and $X_1 \sim N(0, 1)$,  if $X_2<0.5$, else $X_1 \sim N(4,4)$ (black dots). \textbf{Top left:} The crosses are permutations of $X_1$. For $X_2<0.5$, the permutation extrapolates. \textbf{Bottom left:} Marginal density of $X_1$. \textbf{Top right:} Permuting $X_1$ within subgroups based on $X_2$ ($X_2<0.5$ and $X_2\geq 0.5$) reduces extrapolation. \textbf{Bottom right:} Densities of $X_1$ conditional on the subgroups.}\label{fig:split-density}
\end{figure}

%
We suggest approaching the dependent feature problem by constructing an interpretable grouping $G_j$ such that the feature of interest $X_j$ becomes less dependent on remaining features $X_{-j}$ within each subgroup.
In the best case the features become independent: $(X_j \perp X_{-j}) | G_j$.
Assuming that we find a grouping in which $(X_j \perp X_{-j}) | G_j$ holds, sampling from the group-wise marginal distribution removes extrapolation (see Figure \ref{fig:split-density}) and within each group, the samples from the marginal and the conditional distribution would coincide.
Such groupings exist when, for example, the features in $X_{-j}$ are categorical, or when the conditional distribution of $X_j$ only depends on discrete changes in features $X_{-j}$.
Such a grouping would consequently enable (1) the application of standard PFI and PDP within each group \textit{without extrapolation} and (2) sampling from the global conditional distribution $P(X_j|X_{-j})$ using group-wise permutation and aggregation.
With our approach we exploit these properties to derive both a group-wise marginal interpretation and, for the PFI, a global conditional interpretation.
Even when such a discrete grouping does not exist, e.g., when the true dependence is linear, the cs-permutation reduces extrapolation, see Figure~\ref{fig:gaussian}.
Moreover, an accurate interpretation requires the groupings to be human-intelligible.
We can gain insight into how the model behaves within specific subgroups which is not possible with approaches that directly sample $X_j$ conditional on all features $X_{-j}$  \citep{candes2018panning,strobl2008conditional,aas2019explaining,fisher2019all,watson2019testing}.\\

For our approach, any algorithm can be used that splits the data in $X_{-j}$ so that the distribution of $X_j$ becomes more homogeneous within a group and more heterogeneous between groups.
We consider decision tree algorithms for this task, which predict $X_j$ based on splits in $X_{-j}$.
Decision tree algorithms directly or indirectly optimize splits for heterogeneity of some aspects of the distribution of $X_j$ in the splits.
The partitions in a decision tree can be described by decision rules that lead to that terminal leaf.
We leverage this partitioning to construct an interpretable grouping $\mathcal{G}^k_j$ based on random variable $G_j$ for a specific feature $X_j$.
The new variable can be calculated by assigning every observation the indicator of the partition that it lies in (meaning for observation $i$ with $x_{-j}^{(i)} \in \mathcal{G}^k_j$ the group variable's value is defined as $g_j^{(i)}:=k$).

\textbf{Transformation trees  (trtr)} \citep{hothorn2017transformation} are able to model the conditional distribution of a variable.
This approach partitions the feature space so that the distribution of the target (here $X_j$) within the resulting subgroups $\mathcal{G}^k_j$ is homogeneous, which means that the group-wise parameterization of the modeled distribution is independent of $X_{-j}$.
Transformation trees directly model the target's distribution $\P(X_j \leq x) = F_Z(h(x))$, where $F_Z$ is the chosen (cumulative) distribution function and $h$ a monotone increasing transformation function (hence the name transformation trees).
The transformation function is defined as $\mathbf{a}(y)^T \boldsymbol\theta$ where $\mathbf{a}:\Xspace_j \mapsto \R^k$ is a basis function of polynomials or splines.
The task of estimating the distribution is reduced to estimating $\boldsymbol{\theta}$, and the trees are split based on hypothesis tests for differences in $\boldsymbol{\theta}$ given $X_{-j}$, and therefore differences in the distribution of $X_j$.
For more detailed explanations of transformation trees please refer to \cite{hothorn2017transformation}.

In contrast, a simpler approach would be to use \textbf{classification and regression trees (CART)} \citep{cart}, which, for regression, minimizes the variance within nodes, effectively finding partitions with different means in the distribution of $X_j$.
However, CART's split criterion only considers differences in the expectation of the distribution of $X_j$ given $X_{-j}$: $\E[X_j|X_{-j}]$.
This means CART could only make $X_j$ and $X_{-j}$ independent if the distribution of $X_j$ only depends in its expectation on $X_{-j}$ (and if the dependence can be modeled by partitioning the data).
Any differences in higher moments of the distribution of $X_j$ such as the variance of $X_j | X_{-j}$ cannot be detected.

We evaluated both trtr which are theoretically well equipped for splitting distributions and CART, which are established and well-studied.
For the remainder of this paper, we have set the default minimum number of observations in a node to 30 for both approaches.
For the transformation trees, we used the Normal distribution as target distribution and we used Bernstein polynomials of degree five for the transformation function.
Higher-order polynomials do not seem to increase model fit further \citep{hothorn2018top}.

We denote the subgroups by $\mathcal{G}^k_j \subset \R^{p-1}$, where $k \in \{1,\ldots,K_j\}$ is the $k$-th subgroup for feature $j$, with $K_j$ groups in total for the $j$-th feature.
The subgroups  per feature are disjoint: $\mathcal{G}^l_j \cap \mathcal{G}^k_j = \emptyset, \forall l \neq k$ and $\bigcup_{k=1}^K \mathcal{G}^k_j = \R^{p-1}$.
Let $(\mathbf{y}^k_j, \mathbf{x}^k_j)$ be a subset of $(\mathbf{y}, \mathbf{x})$ that refers to the data subset belonging to the subgroup $\mathcal{G}^k_j$.
Each subgroup can be described by the decision path that leads to the respective terminal node.

\subsection{Remarks}

\subsubsection{Continuous Dependencies}
\begin{figure}
  \includegraphics[width=\columnwidth]{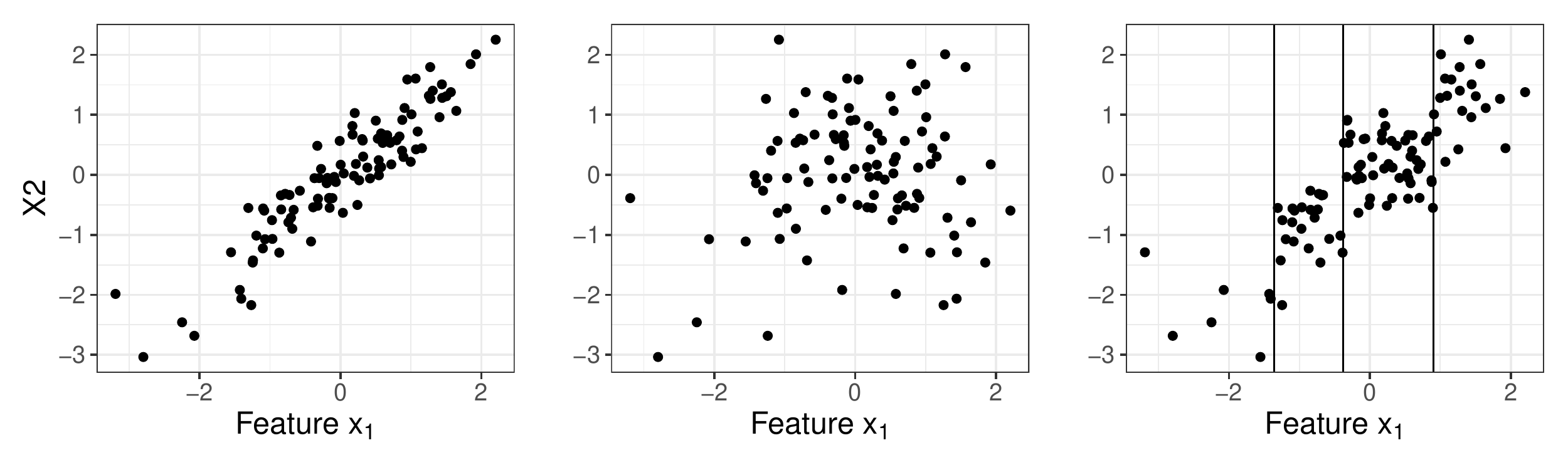} \caption{\textbf{Left:} Simulation of features $X_1 \sim N(0,1)$ and $X_2 \sim N(0,1)$ with a covariance of 0.9.  \textbf{Middle:} Unconditional permutation extrapolates strongly. \textbf{Right:} Permuting on partitions found by CART (predicting $X_2$ from $X_1$) has greatly reduces extrapolation, but cannot get rid of it completely. $x_1$ and $x_2$ remain correlated in the partitions.
  }\label{fig:gaussian}
\end{figure}

For conditional independence $X_j \perp X_{-j} | G_j^k$ to hold, the chosen decision tree approach has to capture the (potentially complex) dependencies between $X_j$ and $X_{-j}$.
CART can only capture differences in the expected value of $X_j|X_{-j}$ but are insensitive to changes in, for example, the variance.
Transformation trees are in principle agnostic to the specified distribution and the default transformation family of distributions is very general, as empirical results suggest \citep{hothorn2017transformation}.
However, the approach is based on the assumption that the dependence can be modeled with a discrete grouping.
For example, in the case of linear Gaussian dependencies, the corresponding optimal variable would be linear Gaussian itself, and would be in conflict with our proposed interpretable grouping approach.
Even in these settings the approach allows an approximation of the conditional distribution.
In the case of simple linear Gaussian dependencies, partitioning the feature space will still \textbf{reduce extrapolation}.
But we never get rid of it completely, unless there are only individual data points left in each partition, see Figure~\ref{fig:gaussian}.
\subsubsection{Sparse Subgroups}
Fewer subgroups are generally desirable for two reasons:
(1) for a good approximation of the marginal distribution within a subgroup, a sufficient number of observations per group is required, which might lead to fewer subgroups, and
(2) a large number of subgroups leads to more complex groups, which reduces their human-intelligibility and therefore forfeits the added value of the local, subgroup-wise interpretations.
As we rely on decision trees, we can adjust the granularity of the grouping using hyperparameters such as the maximum tree depth.
By controlling the maximum tree depth, we can control the trade-off between the depth of the tree (and hence its interpretability) and the homogeneity of the distribution within the subgroups.

\subsection{Conditional Subgroup Permutation Feature Importance (cs-PFI)}

We estimate the cs-PFI of feature $X_j$ within a subgroup $\mathcal{G}^k_j$ as:
\begin{equation}
  PFI_j^k =  \frac{1}{n_k}\sum_{i: \xi \in \mathcal{G}^k_j} \left(\frac{1}{M}\sum_{m=1}^M L(\yi, \fh(\tilde{x}^{m(i)}_{j}, \xi_{-j})) - L(\yi, \fh(\xi))\right),
  \label{eq:cspfi}
\end{equation}
where $\tilde{x}^{m(i)}_{j}$ refers to a feature value obtained from the $m$-th permutation of $x_j$ within the subgroup $k_j$.
This estimation is exactly the same as the marginal PFI (Equation~\ref{eq:pfi}), except that it only includes observations from the given subgroup.
Algorithm~\ref{algo:cspfi} describes the estimation of the cs-PFIs for a given feature on unseen data.

\begin{algorithm}
\label{algo:cspfi}
\caption{Estimate cs-PFI}
        \KwInput{Model $f$; data $\mathcal{D}_{train}$, $\mathcal{D}_{test}$; loss $L$; feature $j$; no. permutations $M$}
        Train tree $T_j$ with target $X_j$ and features $X_{-j}$ using $\mathcal{D}_{train}$\;
        Compute subgroups $\mathcal{G}^k_j$ for $\mathcal{D}_{test}$ based on terminal nodes of $T_j$, $k \in \{1,\ldots,K_j\}$\;
        \For{$k \in \{1,\ldots,K_j\}$}{
           $L_{orig} := \frac{1}{n_k}\sum_{i: \xv^{(i)}\in \mathcal{G}^k_j} L(\yi, \fh(\xv^{(i)}))$\;
            \For{$m \in \{1,\ldots,M\}$}{
              Generate $\tilde{\xv}^{m}_j$ by permuting feature values $\xv_j$ within subgroup $\mathcal{G}^k_j$ \;
             $L_{perm}^m := \frac{1}{n_k}\sum_{i: \xv^{(i)} \in \mathcal{G}^k_j} L(\yi, \fh(\tilde{x}^{m(i)}_{j}, \xv^{(i)}_{-j}))$\;
          }
            $\text{cs-PFI}_j^k = \frac{1}{M}\sum_{m=1}^M L_{perm}^m - L_{orig}$\;
        }
        $\text{cs-PFI}_j = \frac{1}{n}\sum_{k=1}^{K_j}n^k PFI^k_j$\;
\end{algorithm}

The algorithm has two outcomes:
We get local importance values for feature $X_j$ for each subgroup ($\text{cs-PFI}^k_j$; Algorithm~\ref{algo:cspfi}, line 8) and a global conditional feature importance ($\text{cs-PFI}_j$; Algorithm~\ref{algo:cspfi}, line 9).
The latter is equivalent to the weighted average of subgroup importances regarding the number of observations within each subgroup (see proof in Appendix~\ref{app:decompose-cpfi}).
$$\text{cs-PFI}_j = \frac{1}{n}\sum_{k=1}^{K_j} n^k PFI^k_j$$

The cs-PFIs needs the same amount of model evaluations as the PFI ($O(nM)$).
On top of that comes the cost for training the respective decision trees and making predictions to assign a subgroup to each observation.

\begin{theorem}
  When feature $X_j$ is independent of features $X_{-j}$ for a given dataset $\D$, each $\text{cs-PFI}_j^k$ has the same expectation as the marginal PFI, and an $n/n_k$-times larger variance, where $n$ and $n_k$ are the number of observations in the data and the subgroup $\mathcal{G}^k_j$.
  \label{th:cspfi}
\end{theorem}
The proof of Theorem~\ref{th:cspfi} is shown in Appendix~\ref{app:exp-var-cspfi}.
Theorem~\ref{th:cspfi} has the practical implication that even in the case of applying cs-PFI to an independent feature, we will retrieve the marginal PFI, and not introduce any problematic interpretations.
Equivalence in expectation and higher variance under the independence of $X_j$ and $X_{-j}$ holds true even if the partitions $\mathcal{G}_j^k$ would be randomly chosen.
Theorem~\ref{th:cspfi} has further consequences regarding overfitting:
Assuming a node has already reached independence between $X_j$ and $X_{-j}$, then further splitting the tree based on noise will not change the expected cs-PFIs.

\subsection{Conditional Subgroup Partial Dependence Plots (cs-PDPs)}

\begin{figure}
    \centering
    \includegraphics[width = 0.8\textwidth]{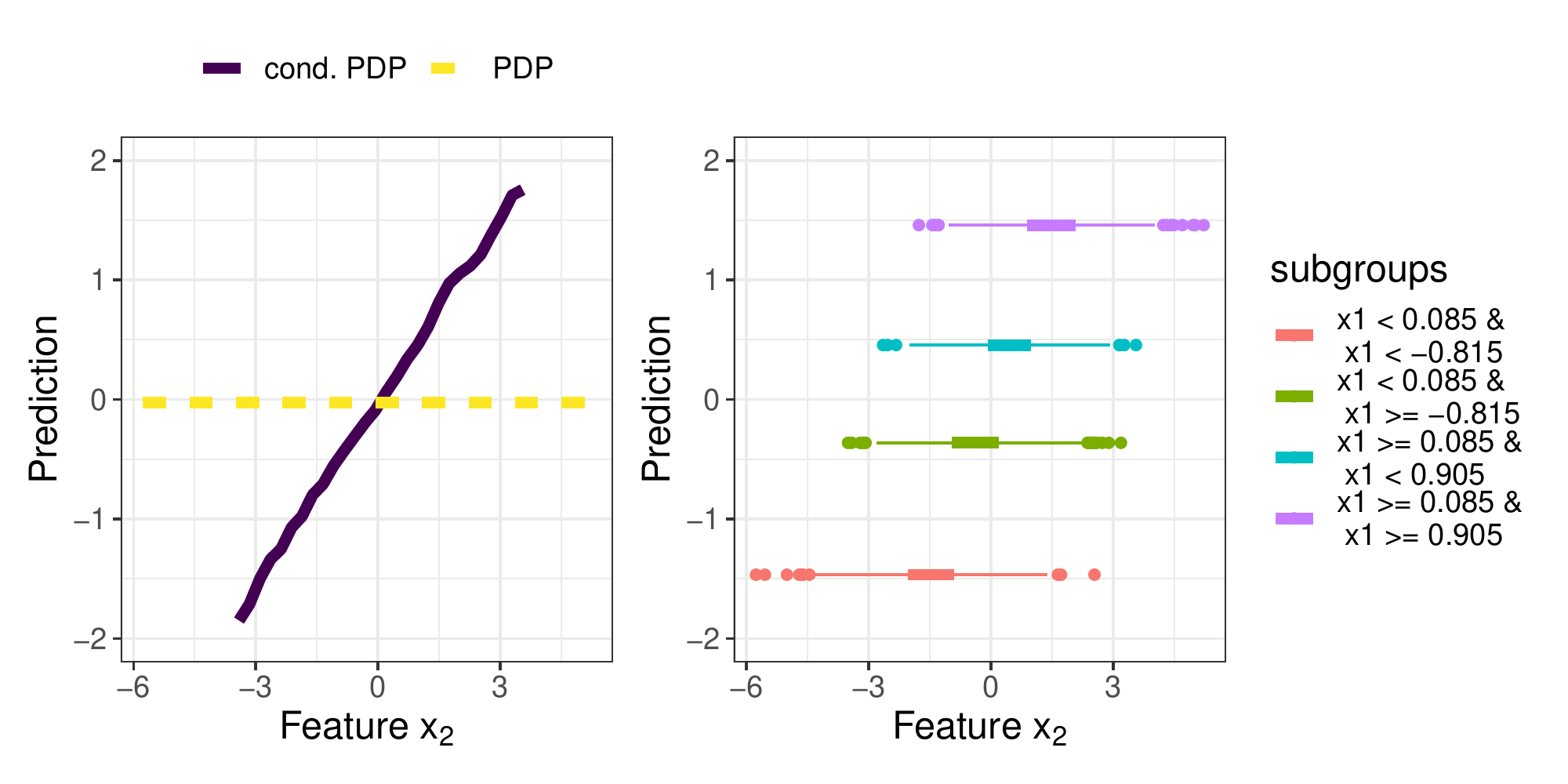}
    \caption{We simulated a linear model of $y = x_1 + \epsilon$ with $\epsilon \sim N(0,1)$ and an additional feature $X_2$ which is correlated with $X1$ ($\approx 0.72$). The conditional PDP (left) gives the false impression that $X_2$ has an influence on the target. The cs-PDPs help in this regard, as the effects due to $X_1$ (changes in intercept) are clearly separated from the effect that $X_2$ has on the target (slope of the cs-PDPs), which is zero. Unlike the  marginal PDP, the cs-PDPs reveals that for increasing $X_2$ we expect that the prediction increases due to the correlation between $X_1$ and $X_2$.}
    \label{fig:mplot-fail-again}
\end{figure}

\begin{figure}
\centering
  \includegraphics[width=0.8\textwidth]{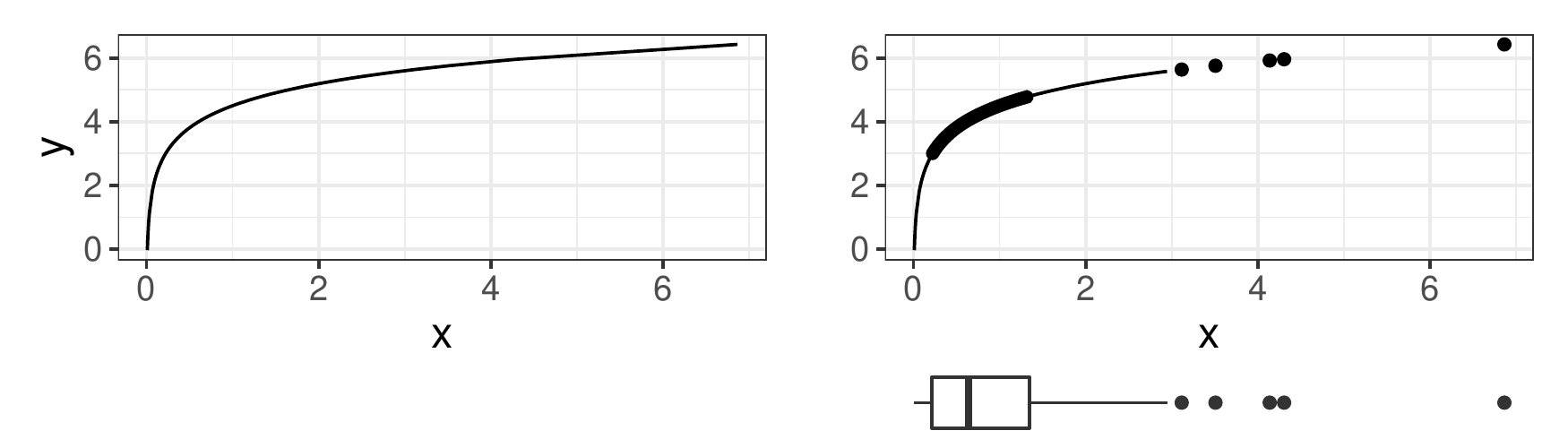} \caption{\textbf{Left:} Marginal PDP. \textbf{Bottom right:} Boxplot showing the distribution of feature $X$. \textbf{Top right:} PDP with boxplot-like emphasis. In the $x$-range, the PDP is drawn from $\pm1.58 \cdot IQR/\sqrt{n}$, , where $IQR$ is the range between the $25\%$ and $75\%$ quantile. If this range exceeds  $[min(x_j),max(x_j)]$, the PDP is capped. Outliers are drawn as points. The PDP is bold between the $25\%$ and $75\%$ quantiles.}
\label{fig:pdp-boxplot}
\end{figure}

The conditional PDP has a different interpretation than the marginal PDP, as the motivating example in Figure \ref{fig:mplot-fail} showed:
The conditional PDP can be interpreted as the effect of a feature on the prediction, given that all other features would change according to the joint distribution.
This violates a desirable property that the effect of features that were not used by the model should have a zero effect curve.
This poses a dilemna for dependent features:
Either extrapolate using the marginal PDP, or use the conditional PDP with undesirable properties for interpretation.
Our proposed cs-PDPs reduces extrapolation while allowing a marginal interpretation \textit{within} each subgroup.
We compute the $\text{cs-PDP}_j^k$ for each subgroup $\mathcal{G}^k_j$ using the marginal PDP formula in Equation \ref{eq:pdp}.
$$\text{cs-PDP}^k_j (x) = \frac{1}{n^k}\sum_{i:x^{(i)}\in \mathcal{G}^k_j} \fh(x, x^{(i)}_{-j})$$
This results in multiple cs-PDPs per feature, which can be displayed together in the same plot as in Figure \ref{fig:bike-effects-temp}.
As shown in Figure~\ref{fig:mplot-fail-again}, even features that do not contribute to the prediction at all can have a conditional PDP different from zero.
We therefore argue that an aggregation of the cs-PDPs to the conditional PDP is not meaningful for model interpretation, and we suggest to plot the group-wise curves.
For the visualization of the cs-PDPs, we suggest to plot the PDPs similar to boxplots, where the dense center quartiles are indicated with a bold line (see Figure \ref{fig:pdp-boxplot}).
We restrict each $\text{cs-PDP}^k_j$ to the interval $[min(\xv_j),max(\xv_j)], \text{ with } \xv_j = (x_j^{(1)}, \cdots, x_j^{(n_j^k)})$.

%
%

Equivalently to PFI, the subgroup PDPs approximate the true marginal PDP even if the features are independent.
\begin{theorem}
  When feature $X_j$ is independent of features $X_{-j}$ for a given dataset $\D$, each $\text{cs-PDP}_j^k$ has the same expectation as the marginal PDP, and an $n/n_k$-times larger variance, where $n$ and $n_k$ are the number of observations in the data and the subgroup $\mathcal{G}^k_j$.
  \label{th:cspdp}
\end{theorem}
The proof of Theorem~\ref{th:cspdp} is shown in Appendix~\ref{app:exp-var-cspdp}.
Theorem~\ref{th:cspdp} has the same practical implications as Theorem~\ref{th:cspfi}: Even if the features are independent, we will, in expectation, get the marginal PDPs.
And when trees are grown deeper than needed, in expectation the cs-PDPs will yield the same curve.

Both the PDP and the set of cs-PDPs need $O(nM)$ evaluations, since $\sum_{k=1}^{K_j} n^k = n$ (and worst case $O(n^2)$ if evaluated at each $\xij$ value).
Again, there is an additional cost for training the respective decision trees and making predictions.

\section{Training Conditional Sampling Approaches}
\label{sec:perturbation-training}
To ensure that sampling approaches are not overfitting, we suggest to separate training and sampling, where training covers all estimation steps that involve data.
For this purpose, we refer to the training data with $\D_{train}$ and to the data for importance computation with $\D_{test}$.
This section both describes how we compared the sampling approaches in the following chapters and serves as a general recommendation for how to use the sampling approaches.

For our cs-permutation, we trained the CART / transformation trees on $\D_{train}$ and permuted $X_j$ of $\D_{test}$ within the terminal nodes of the tree.
For CVIRF \citep{strobl2008conditional,debeer2020conditional}, which is specific to random forests, we trained the random forest on $\D_{train}$ to predict the target $y$  and permuted $X_j$ of $\D_{test}$ within the terminal nodes.
For Model-X knockoffs \citep{candes2018panning}, we fitted the second-order knockoffs on $\D_{train}$ and replaced $X_j$ in $\D_{test}$ with its knockoffs.
For the imputation approach \citep{fisher2019all}, we trained a random forest on $\D_{train}$ to predict $X_j$ from $X_{-j}$,  and replaced values of $X_j$ in $\D_{test}$ with their random forest predictions plus a random residual.
For the interval-based sampling \citep{apley2016visualizing}, we computed quantiles of $X_j$ using $\D_{train}$ and perturbed $X_j$ in $\D_{test}$ by moving each observation once to the left and once to the right border of the respective intervals.
The marginal permutation (PFI, PDP) required no training, we permuted (i.e., shuffled) the feature $X_j$ in $\D_{test}$.

\section{Conditional PFI Ground Truth Simulation}
\label{sec:cspfi-eval}

We compared our cs-PFI approach using CART (tree cart) and transformation trees (tree trtr), CVIRF \citep{strobl2008conditional,debeer2020conditional}, Model-X knockoffs (ko) \citep{candes2018panning} and the imputation approach (impute rf) \citep{fisher2019all} in ground truth simulations.
We simulated the following data-generating process: $y^{(i)} = f(\xi) =  \xi_1 \cdot \xi_2 + \sum_{j=1}^{10} x_j^{(i)} + \epsilon^{(i)}$, where $\epsilon^{(i)} \sim N(0, \sigma_{\epsilon})$.
All features, except feature $X_1$ followed a Gaussian distribution: $X_j \sim N(0, 1)$.
Feature $X_1$ was simulated as a function of the other features plus noise: $x_1^{(i)} = g(x_{-1}^{(i)}) + \epsilon_x$.
We simulated the following scenarios by changing $g$ and $\epsilon_x$:
\begin{itemize}
    \item In the \textbf{independent} scenario, $X_1$ did not depend on any feature: $g(\xi_{-1}) = 0$, $\epsilon_x \sim N(0,1)$. This scenario served as a test how the different conditional PFI approaches handle the edge case of independence.
    \item The \textbf{linear} scenario introduces a strong correlation of $X_1$ with feature $X_2$: $g(\xi_{-1}) = \xi_2$, $\epsilon_x \sim N(0,1)$.
    \item In the \textbf{non-linear} scenario, we simulated $X_1$ as a non-linear function of multiple features: $g(\xi_{-1}) = 3 \cdot \mathds{1}(\xi_2 > 0) - 3 \cdot \mathds{1}(\xi_2 \leq 0) \cdot \mathds{1}(\xi_3 > 0)$. Here also the variance of $\epsilon_x \sim N(0, \sigma_x)$ is a function of $x$: $\sigma_x(\xi) =  \mathds{1}(\xi_2 > 0) + 2 \cdot \mathds{1}(\xi \leq 0) \cdot \mathds{1}(\xi_3 > 0) + 5 \cdot \mathds{1}(\xi_2 \leq 0) \cdot \mathds{1}(\xi_3 \leq 0)$.
    \item For the \textbf{multiple linear dependencies} scenario, we chose $X_1$ to depend on many features: $g(\xi_{-1}) = \sum_{j=2}^{10} \xij$, $\epsilon_x \sim N(0,5)$.
\end{itemize}

For each scenario, we varied the number of sampled data points $n \in \{300, 3000\}$ and the number of features $p \in \{9, 90\}$.
To \enquote{train} each of the cPFI methods, we used $2/3 \cdot n$ (200 or 2000) data points and the rest (100 / 1000) to compute the cPFI.
The experiment was repeated 1000 times.
We examined two settings.
\begin{itemize}
\item In setting (I), we assumed that the model recovered the true model $\fh = f$.
\item In setting (II), we trained a random forest with 100 trees \citep{breiman2001random}.
\end{itemize}

In both settings, the true conditional distribution of $X_1$ given the remaining features is known (function $g$ and error distribution is known).
Therefore we can compute the ground truth conditional PFI, as defined in Equation~\ref{eq:pfi}.
We generated the samples of $X_1$ according to $g$ to get the $\tilde{X}_1$ values and compute the increase in loss.
The conditonal PFIs differed in settings (I) and (II) since in (I) we used the true $f$, and in (II) the trained random forest $\fh$.

\subsection{Conditional PFI Ground Truth Results}

\begin{figure}
  \includegraphics[width=\columnwidth]{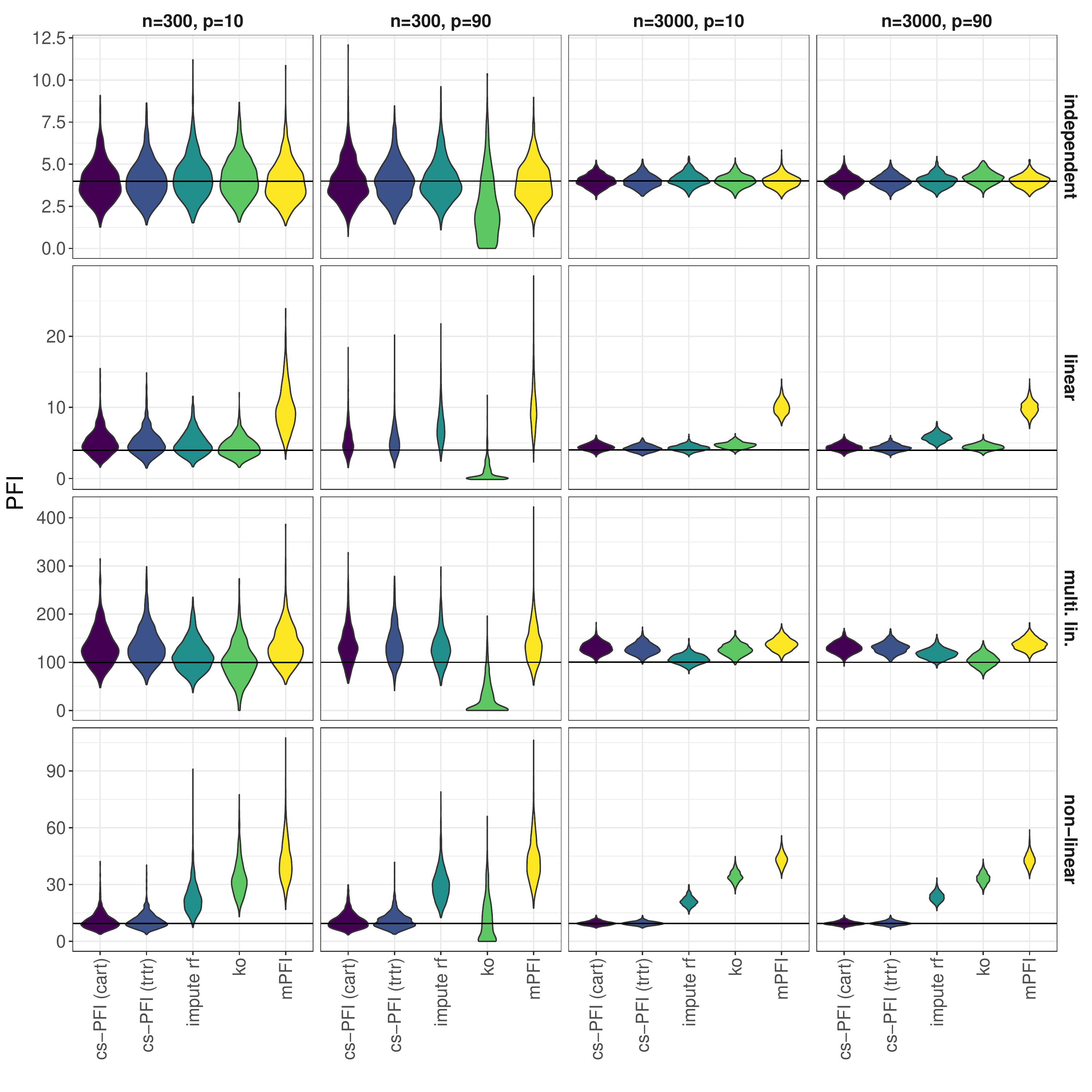} \caption{Setting (I) comparing various conditional PFI approaches on the true model against the true conditional PFI (horizontal line) based on the data generating process.}
  \label{fig:true-importance-all-ex1}
\end{figure}

For setting (I), the mean squared errors between the estimated conditional PFIs and the ground truth are displayed in Table~\ref{tab:mses-ex1}, and the distributions of conditional PFI estimates in Figure~\ref{fig:true-importance-all-ex1}.
In the \textit{independent scenario}, where conditional and marginal PFI are equal, all methods performed equally well, except in the low $n$, high $p$ scenario, where the knockoffs sometimes failed.
As expected, the variance was higher for all methods when $n=300$.
In the \textit{linear scenario}, the marginal PFI was clearly different from the conditional PFI.
There was no clear best performing conditional PFI approach, as the results differ depending on training size $n$ and number of features $p$.
For low $n$ and low $p$, knockoffs performed best.
For high $p$, regardless of $n$, the cs-permutation approaches worked best, which might be due to the feature selection mechanism inherent to trees.
The \textit{multiple linear dependencies scenario} was the only scenario in which the cs-PFI approach was consistently outperformed by the other methods.
Decision trees already need multiple splits for recovering linear relationships, and in this scenario, multiple linear relationships had to be recovered.
Imputation with random forest worked well when multiple linear dependencies are present.
For knockoffs, the results were mixed.
As expected, the cs-PFI approach worked well in the \textit{non-linear scenario}, and outperformed all other approaches.
Knockoffs and imputation with random forests both overestimated the conditional PFI (except for knockoffs for $n=300$ and $p=90$). 
In addition to this bias, they had a larger variance compared to the cs-PFI approaches.

Generally, the transformation trees performed equal to or outperformed CART across all scenarios, except for the multiple linear dependencies scenario.
Our cs-PFI approaches worked well in all scenarios, except when multiple (linear) dependencies were present.
Even for a single linear dependence, the cs-PFI approaches were on par with knockoffs and imputation, and clearly outperformed both when the relationship was more complex.

\begin{table}

\caption{\label{tab:mses-ex1}MSE comparing estimated and true conditional PFI (scenario I). Legend: impute rf: Imputation with a random forest, ko: Model-X knockoffs, mPFI: (marginal) PFI, tree cart: cs-permutation based on CART, tree trtr: cs-permutation based on transformation trees.}
\centering
\begin{tabular}[t]{lrrrrr}
\toprule
setting & cs-PFI (cart) & cs-PFI (trtr) & impute rf & ko & mPFI\\
\midrule
\addlinespace[0.3em]
\multicolumn{6}{l}{\textbf{independent}}\\
\hspace{1em}n=300, p=10 & 1.33 & 1.35 & 1.67 & 1.47 & 1.39\\
\hspace{1em}n=300, p=90 & 1.50 & 1.29 & 1.46 & 5.81 & 1.31\\
\hspace{1em}n=3000, p=10 & 0.14 & 0.15 & 0.16 & 0.13 & 0.15\\
\hspace{1em}n=3000, p=90 & 0.15 & 0.14 & 0.14 & 0.18 & 0.13\\
\addlinespace[0.3em]
\multicolumn{6}{l}{\textbf{linear}}\\
\hspace{1em}n=300, p=10 & 4.62 & 4.30 & 3.64 & 2.03 & 44.83\\
\hspace{1em}n=300, p=90 & 5.55 & 5.26 & 17.53 & 11.63 & 45.36\\
\hspace{1em}n=3000, p=10 & 0.40 & 0.26 & 0.26 & 0.63 & 37.40\\
\hspace{1em}n=3000, p=90 & 0.45 & 0.31 & 3.55 & 0.38 & 36.32\\
\addlinespace[0.3em]
\multicolumn{6}{l}{\textbf{multi. lin.}}\\
\hspace{1em}n=300, p=10 & 2443.67 & 2623.54 & 1276.41 & 1583.69 & 2739.83\\
\hspace{1em}n=300, p=90 & 2574.54 & 2896.47 & 2141.01 & 6607.73 & 2988.68\\
\hspace{1em}n=3000, p=10 & 1031.83 & 900.68 & 140.98 & 810.78 & 1548.37\\
\hspace{1em}n=3000, p=90 & 1075.95 & 1041.10 & 438.25 & 185.13 & 1599.59\\
\addlinespace[0.3em]
\multicolumn{6}{l}{\textbf{non-linear}}\\
\hspace{1em}n=300, p=10 & 22.00 & 17.76 & 265.73 & 668.34 & 1204.17\\
\hspace{1em}n=300, p=90 & 19.99 & 19.81 & 504.53 & 131.77 & 1248.74\\
\hspace{1em}n=3000, p=10 & 1.18 & 1.00 & 144.77 & 626.80 & 1156.32\\
\hspace{1em}n=3000, p=90 & 1.17 & 1.13 & 206.01 & 579.02 & 1136.83\\
\bottomrule
\end{tabular}
\end{table}

In setting (II), a random forest was analyzed, which allowed us to include the conditional variable importance for random forests (CVIRF) by \cite{strobl2008conditional,debeer2020conditional} in the benchmark.
The MSEs are displayed in Appendix~\ref{app:simulation-rf}, Table~\ref{tab:mses-ex2}, and the distribution of conditional PFI estimates in Appendix~\ref{app:simulation-rf} in Figure~\ref{fig:true-importance-all-ex2}.
The results for all other approaches are comparable to setting (I).
For the low $n$ settings, CVIRF worked as well as the other approaches in the \textit{independent scenario}.
It outperformed the other approaches in the \textit{linear scenario} and the \textit{multiple linear scenario} (when $n$ was small).
The CVIRF approach consistently underestimated the conditional PFI in all scenarios with high $n$, even in the \textit{independent scenario}.
Therefore, we would recommend to analyze the conditional PFI for random forests using cs-PFI for lower dimensional dependence structures, and imputation for multiple (linear) dependencies.

\section{Trading Interpretability for Accuracy}

\begin{figure}
  \includegraphics[width=\columnwidth]{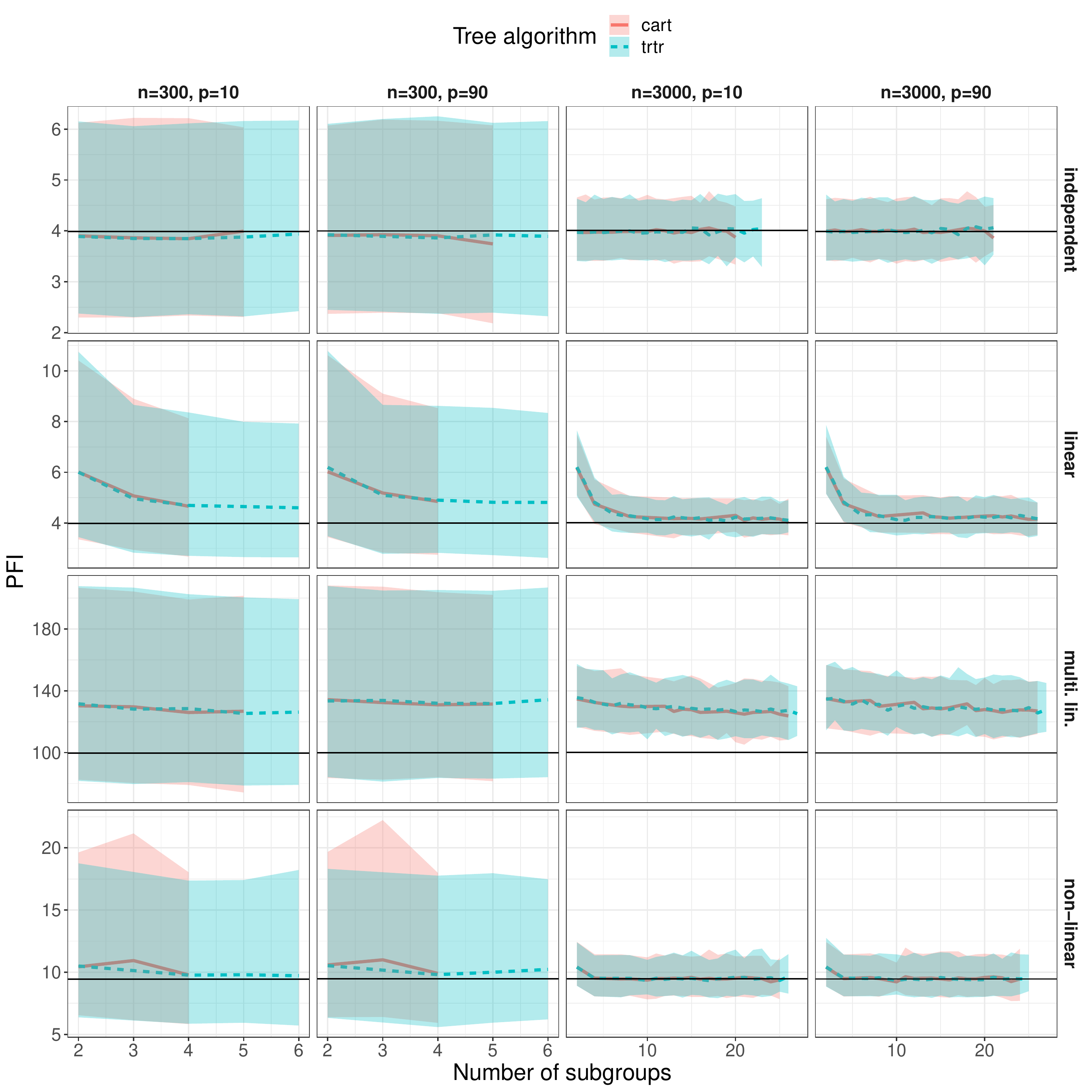} \caption{Conditional PFI estimate using cs-PFI (\textbf{cart} / \textbf{tr}ansformation \textbf{tr}ee) with increasing number of subgroups (simulation scenario I). Displayed is the median PFI over 1000 repetitions along with the 5\% and 95\% quartiles.}
  \label{fig:true-importance-trees}
\end{figure}

In an additional experiment, we examined the trade-off between the depth of the trees and the accuracy with which we recover the true conditional PFI.
For scenario (I), we trained decision trees with different maximal depths (from 1 to 10) and analyzed how the resulting number of subgroups influenced the conditional PFI estimate.
The experiment was repeated 1000 times.
Figure \ref{fig:true-importance-trees} shows that the deeper the transformation trees (and the more subgroups), the better the true conditional PFI was approximated.
 The plot also shows that no overfitting occurred, which is in line with theoretical considerations in Theorem~\ref{th:cspfi}.

\section{Data Fidelity Evaluation}
\label{sec:eval-data-fidelity}

PDP and PFI work by data intervention, prediction, and subsequent aggregation \citep{scholbeck2019sampling}.
Based on data $\D$, the intervention creates a new data set.
In order to compare different conditional sampling approaches, we define a measure of data fidelity to quantify the ability to preserve the joint distribution under intervention.
Failing to preserve the joint distribution leads to extrapolation when features are dependent.
Model-X knockoffs, for example, are directly motivated by preserving the joint distribution, while others, such as accumulated local effect plots do so more implicitly.

Data fidelity is the degree to which a sample $\tilde{X}_j$ of feature $X_j$ preserves the joint distribution, that is, the degree to which $(\tilde{X}_j,X_{-j}) \sim (X_j, X_{-j})$
In theory, any measure that compares two multivariate distributions can be used to compute the data fidelity.
In practice, however, the joint distribution is unknown, which makes measures such as the Kullback-Leibler divergence impractical.
We are dealing with two samples, one data set without and one with intervention.

In this classic two-sample test-scenario, the maximum mean discrepancy (MMD) can be used to compare whether two samples come from the same distribution \citep{fortet1953convergence,smola2007hilbert,gretton2007kernel,gretton2012kernel}.
The empirical MMD is defined as:
\begin{equation}
  \text{MMD}(\D, \tilde{\D}) = \frac{1}{n^2}\sum_{x,z\in \D} k(x, z) - \frac{2}{nl}\sum_{x\in \D, z\in \tilde{\D}}k(x,z) +  \frac{1}{l^2}\sum_{x,z\in \tilde{D}} k(x,z)
\end{equation}
where $\D = \{x^{(i)}_j,x_{-j}^{(i)}\}_{i=1}^n$ is the original data set and $\tilde{\D} = \{\tilde{x}^{(i)}_j,x_{-j}^{(i)}\}_{i=1}^l$ a data set with perturbed $\xij$.
For both data sets, we scaled numerical features to a mean of zero and a standard deviation of one.
For the kernel $k$ we used the radial basis function kernel for all experiments.
For parameter $\sigma$ of the radial basis function kernel, we chose the median L2-distance between data points which is a common heuristic \citep{gretton2012kernel}.
We measure data fidelity as the negative logarithm of the MMD ($-log(\text{MMD})$) to obtain a more condensed scale where larger values are better.
\begin{definition}[MMD-based Data Fidelity]
  Let $\D$ be a dataset, and $\tilde{D}$ be another dataset from the same distribution, but with an additional intervention. We define the data fidelity as:
  $\text{Data Fidelity} = -log(\text{MMD}(\D, \tilde{\D}))$.
\end{definition}
We evaluated how different sampling strategies (see Table~\ref{tab:perturbations}) affect the data fidelity measure for numerous data sets of the OpenML-CC18 benchmarking suite \citep{bischl2019openml}.
We removed all data sets with 7 or fewer features and data sets with more than 500 features.
See Appendix~\ref{app:data-fidelity} for an overview of the remaining data sets.
For each data set, we removed all categorical features from the analysis, as the underlying sampling strategies of ALE plots and Model-X knockoffs are not well equipped to handle them.
We were foremost interested in two questions:
\begin{itemize}
  \item[A)]  How does cs-permutation compare with other sampling strategies w.r.t. data fidelity? 
  \item[B)]  How do choices of tree algorithm (CART vs. transformation trees) and tree depth parameter affect data fidelity?
\end{itemize}

In each experiment, we selected a data set, randomly sampled a feature and computed the data fidelity of various sampling strategies as described in the pseudo-code in Algorithm~\ref{algo:data-fidelity}.

\begin{algorithm}
\label{algo:data-fidelity}
\caption{Data Fidelity Experiments}
	\KwInput{OpenML-CC18 data sets, sampling strategies}
      \For{data set $\D$ in OpenML-CC18}{
          Remove prediction target from $\D$ (only keep it for CVIRF)\;
          Randomly order features in $\D$\;
          \For{features $j \in \{1, \ldots, 10\}$}{
            \For{repetition $\in \{1, \ldots, 30\}$}{
              Sample $min(10.000, n)$ rows from $\D$\;
              Split sample into $\D_{train}$ (40\%), $\D_{test}$ (30\%) and $\D_{ref}$ (30\%)\;
              \For{each sampling}{
                \enquote{Train} sampling approach using $\D_{train}$\ (e.g., construct subgroups, fit knockoff-generator, ...)\;
              Generate conditional sample $\tilde{X}_j$ for $\D_{test}$\;
              Estimate \text{data fidelity} as $-log(MMD(\D_{ref}, \D_{test}))$
              }
            }
            }
            }
        \Return{Set of data fidelity estimates}
\end{algorithm}

For an unbiased evaluation, we split the data into three pieces: $\D_{train}$ (40\% of rows), $\D_{test}$ (30\% of rows) and $\D_{ref}$ (30\% of rows).
We used $\D_{train}$ to \enquote{train} each sampling method (e.g., train decision trees for cs-permutation, see Section~\ref{sec:perturbation-training}).
We used $\D_{ref}$, which we left unchanged and $\D_{test}$, for which the chosen feature was perturbed to estimate the data fidelity.
For each data set, we chose 10 features at random, for which sampling was applied.
Marginal permutation (which ignores the joint distribution) and "no perturbation" served as lower and upper bounds for data fidelity.
For CVIRF, we only used one tree per random forest as we only compared the general perturbation strategy which is the same for each tree.

We repeated all experiments 30 times with different random seeds and therefore different data splits.
All in all this produced $12210$ results (42 data sets $\times$ (up to) 10 features $\times$ 30 repetitions) per sampling method.
All results are shown in detail in Appendix~\ref{app:data-fidelity} (Figures~\ref{fig:data-mmd-all1},~\ref{fig:data-mmd-all2},~\ref{fig:data-mmd-tree1},~\ref{fig:data-mmd-tree2}).

Since the experiments are repeated across the same data sets and the same features, the data fidelity results are not independent.
Therefore, we used a random intercept model \citep{bryk1992hierarchical} to analyze the differences in data fidelity between different sampling approaches.
The random intercepts were nested for each data set and each feature.
We chose \enquote{Marginal Permutation} as the reference category.
We fitted two random intercept models: One to compare cs-permutation with fully-grown trees (CART, trtr) with other sampling methods and another one to compare different tree depths.

\subsection{Results A) State-of-the-art comparison}
\begin{figure}
  \includegraphics[width=\columnwidth]{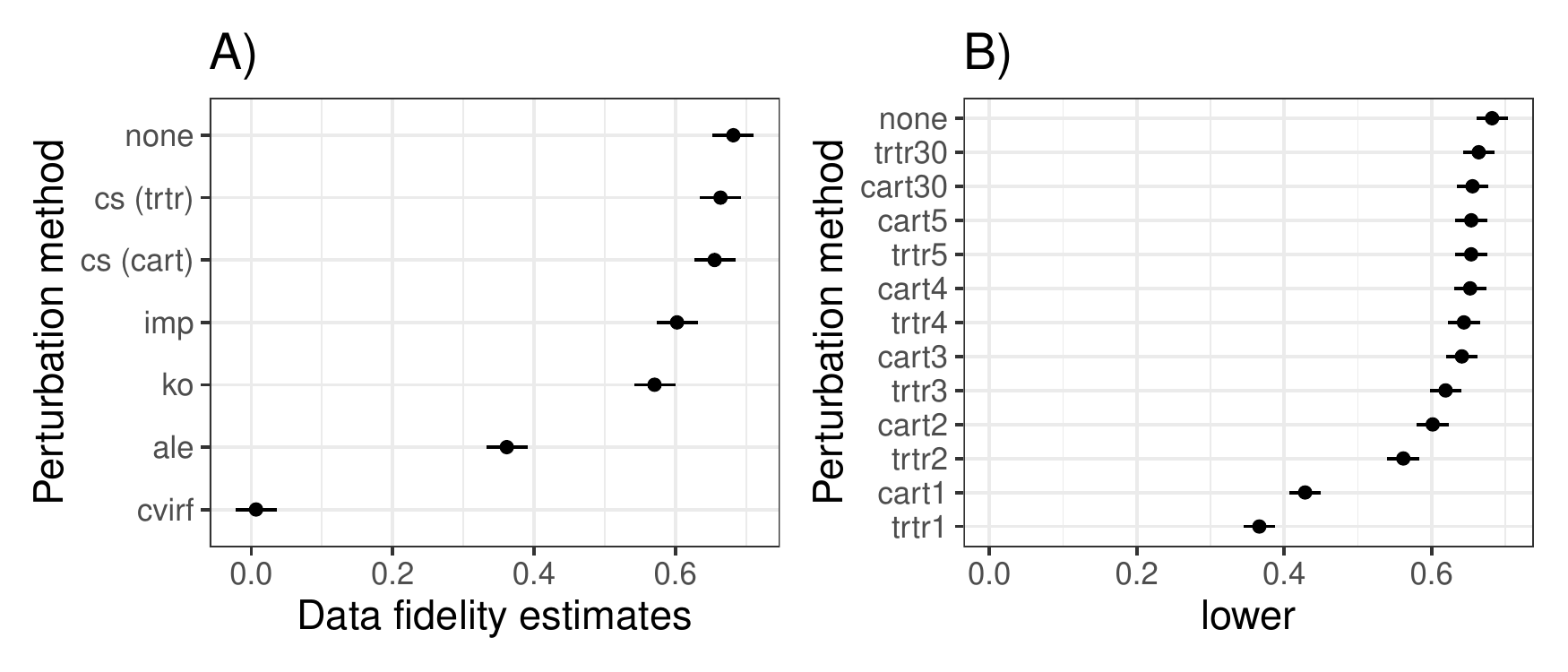} \caption{Linear regression model coefficients and 95\% confidence intervals for the effect of different sampling approaches on data fidelity, with (nested) random effects per data set and feature. \textbf{A)} Comparing different sampling approaches. No perturbation ("none") and permutation ("perm") serve as upper and lower bounds. \textbf{B)} Comparing cs-permutation using either CART or transformation trees and different tree depths (1,2,3,4,5 and 30). Marginal permutation is the reference category.}
\label{fig:data-mmd-cis}
\end{figure}

Figure~\ref{fig:data-mmd-cis} shows the effect estimates of different sampling approaches modeled with a random intercept model.
The results show that cs-permutation performed better than all other methods.
Model-X knockoffs and the imputation approach (with random forests) came in second place and outperformed ALE and CVIRF.
Knockoffs were proposed to preserve the joint distribution, but are based on multivariate Gaussian distribution.
This seems to be too restrictive for the data sets in our experiments.
CVIRF does not have much higher data fidelity than marginal permutation.
However, results for CVIRF must be viewed with caution, since data fidelity regards all features equally -- regardless of their impact on the model prediction.
For example, a feature can be highly correlated with the feature of interest, but might not be used in the random forest.
A more informative experiment for comparing CVIRF can be found in Section~\ref{sec:cspfi-eval}.
Figure~\ref{fig:data-mmd-all1} and Figure~\ref{fig:data-mmd-all2} in Appendix~\ref{app:data-fidelity} show the individual data fidelity results for the OpenML-CC18 data sets.
Not perturbing the feature at all has the highest data fidelity and serves as the upper bound.
The marginal permutation serves as a lower baseline.
For most data sets, cs-permutation has a higher data fidelity compared to all other sampling approaches.
For all the other methods there is at least one data set on which they reach a low data fidelity (e.g., ``semeion", ``qsar-biodeg" for ALE; ``nodel-simulation", ``churn" for imputation; ``jm1", ``pc1" for knockoffs).
In contrast, cs-permutation achieves a consistently high data fidelity on all these data sets.

Additionally, we review the data fidelity rankings of the sampling methods in Table~\ref{tab:ranks}.
The rankings show a similar picture as the random intercept model estimates, except that Model-X knockoffs have a better average ranking than imputation.
This could be the case since on a few data sets (bank-marketing, electricity, see Figure~\ref{fig:data-mmd-all1} in Appendix~\ref{app:data-fidelity}) Model-X knockoffs have a very low data fidelity but on most others a higher model fidelity than the imputation method.

\begin{table}[ht]
\centering
\begin{tabular}{rrrrrrrrr}
  \hline
 & none & cs (trtr) & ko & cs (cart) & imp & ale & perm & cvirf \\ 
  \hline
Mean ranks & 2.50 & 3.51 & 3.70 & 3.76 & 4.25 & 4.61 & 6.82 & 6.84 \\ 
  SD & 0.73 & 0.87 & 1.32 & 0.91 & 1.37 & 2.07 & 1.14 & 1.14 \\ 
   \hline
\end{tabular}
\caption{Mean ranks and their standard deviation based on data fidelity of various perturbation methods over data sets, features and repetitions. \textbf{Legend:} none: No intervention, which serves as upper benchmark. cart30: cs-permutation with CART with maximal depth of 30. trtr30: cs-permutation with transformation trees with maximal depth of 30. imp: Imputation approach. ko: Model-X knockoffs \cite{candes2018panning} . ale: ALE perturbation \cite{apley2016visualizing}. cvirf: Conditional variable importance for random forests \cite{strobl2008conditional}. perm: Unconditional permutation.} 
\label{tab:ranks}
\end{table}

\subsection{Results B) tree configuration}
We included shallow trees with maximum depth parameter from 1 to 5 to analyze the trade-off between tree depth and data fidelity.
We included trees with a maximum depth parameter of 30 (\enquote{fully-grown} trees as this was the software's limit) as an upper bound for each decision tree algorithm.
Figure~\ref{fig:data-mmd-cis} B) shows that the deeper the trees (and the more subgroups), the higher the data fidelity.
This is to be expected, since deeper trees allow for a more fine-grained separation of distributions.
More importantly, we are interested in the trade-off between depth and data fidelity.
Even splitting with a maximum depth of only 1 (two subgroups) strongly improves data fidelity over the simple marginal permutation for most data sets.
A maximum depth of two means another huge average improvement in data fidelity, and already puts cs-permutation on par with knockoffs.
A depth of three to four is almost as good as a maximum depth parameter of 30 and already outperforms all other methods, while still being interpretable due to their shortness.
CART slightly outperforms transformation trees clearly when trees are shallow, which is surprising since transformation trees are, in theory, better equipped to handle changes in the distribution.
Deeply grown transformation trees (max. depth of 30) slightly outperform CART.
Figure~\ref{fig:data-mmd-tree1} and Figure~\ref{fig:data-mmd-tree2} in Appendix~\ref{app:data-fidelity} show data fidelity aggregated by data set.

\section{Model Fidelity}

\label{sec:eval-model-fidelity}

Model fidelity has been defined as how well the predictions of an explanation method approximate the ML model \citep{ribeiro2016should}.
Similar to \cite{szepannek2019much}, we define model fidelity for feature effects as the mean squared error between model prediction and the prediction of the partial function $f_j$ (which depends only on feature $X_j$) defined by the feature effect method, for example $f_j(x) = PDP_j(x)$.
For a given data instance with observed feature value $\xij$, the predicted outcome of, for example, a PDP can be obtained by the value on the y-axis of the PDP at the observed $x_j$ value.\\

\begin{equation}
  \label{eq:mfidel}
   \textstyle \text{Model\_Fidelity}(\fh, f_j)= \frac{1}{n}\sum_{i=1}^n (\fh(x^{(i)}) - f_j(\xij))^2,
\end{equation}

where $f_j$ is a feature effect function such as ALE or PDP.
In order to evaluate ALE plots, they have to be adjusted such that they are on a comparable scale to a PDP \citep{apley2016visualizing}: $f_j^{ALE,adj} = f_j^{ALE} + \frac{1}{n}\sum_{i=1}^n \fh(x^{(i)})$.

We trained random forests (500 trees), linear models and k-nearest neighbours models (k = 7) on various regression data sets (Table \ref{tab:datasets}).
\begin{table}[ht]
\centering
\begin{tabular}{lrrrrrr}
  \hline
 & wine & satellite & wind & space & pollen & quake \\ 
  \hline
No. of rows & 6497 & 6435 & 6574 & 3107 & 3848 & 2178 \\ 
  No. of features & 12 & 37 & 15 &  7 &  6 &  4 \\ 
   \hline
\end{tabular}
\caption{We selected data sets from OpenML \cite{vanschoren2014openml, Casalicchio2017} having 1000 to 8000 instances and a maximum of 50 numerical features. We excluded data sets with categorical features, since ALE cannot handle them.} 
\label{tab:datasets}
\end{table}

70\% of the data were used to train the ML models and the transformation trees / CARTs.
This ensure that results are not over-confident due to overfitting, see also Section~\ref{sec:perturbation-training}.
The remaining 30\% of the data were used to evaluate model fidelity.
For each model and each data set, we measured model fidelity between effect prediction and model prediction (Equation \ref{eq:mfidel}), averaged across observations and features.

\begin{figure}
  \includegraphics[width=\columnwidth]{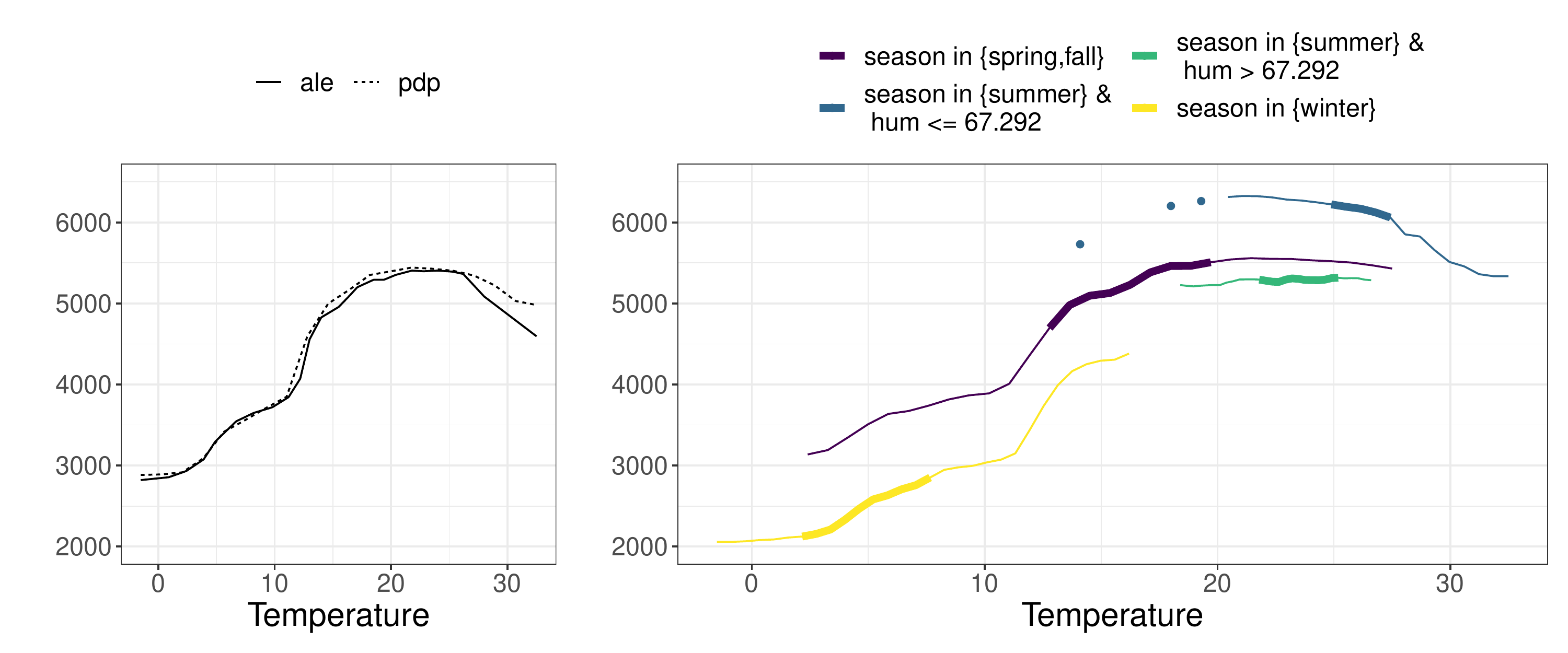} \caption[Effect of temperature on predicted bike rentals]{Effect of temperature on predicted bike rentals. \textbf{Left:} PDP and ALE plot. \textbf{Right:} cs-PDPs for 4 subgroups.}\label{fig:bike-effects-temp}
\end{figure}

Table \ref{tab:model-fidelity} shows that the model fidelity of ALE and PDP is similar, while the cs-PDPs have the best model fidelity.
This is an interesting result since the decision trees for the cs-PDPs are neither based on the model nor on the real target, but solely on the conditional dependence structure of the features.
However, the cs-PDPs have the advantage that we obtain multiple plots.
We did not aggregate the plots to a single conditional PDP, but computed the model fidelity for the PDPs within the subgroups (visualized in Figure~\ref{fig:bike-effects-temp}).
Our cs-PDPs using trees with a maximum depth of 2 have a better model fidelity than using a maximum depth of 1.
We limited the analysis to interpretable conditioning and therefore allowed only trees with a maximum depth of 2, since a tree depth of 3 already means up to 8 subgroups which is already an impractical number of PDPs to have in one plot.
CART sometimes beats trtr (e.g., on the ``satellite" data set) but sometimes trtr has a lower loss (e.g., on the ``wind" data set).
 Using different models (knn or linear model) produced similar results, see Appendix~\ref{sec:model-fid-extra}.

\begin{table}[ht]
\centering
\begin{tabular}{rrrrrrr}
  \hline
 & pollen & quake & satellite & space & wind & wine \\ 
  \hline
PDP & 9.61 & 0.04 & 4.80 & 0.03 & 44.84 & 0.75 \\ 
  ALE & 9.91 & 0.04 & 4.81 & 0.03 & 44.83 & 0.75 \\ 
  cs-PDP trtr1 & 8.44 & 0.04 & 4.49 & 0.03 & 29.96 & 0.71 \\ 
  cs-PDP cart1 & 8.44 & 0.04 & 3.71 & 0.03 & 31.38 & 0.73 \\ 
  cs-PDP trtr2 & 8.17 & 0.04 & 3.25 & 0.03 & 26.56 & 0.70 \\ 
  cs-PDP cart2 & 8.29 & 0.04 & 3.05 & 0.03 & 25.96 & 0.71 \\ 
   \hline
\end{tabular}
\caption{Median model fidelity averaged over features in a random forest for various data sets. The cPDPs always had a lower loss (i.e. higher model fidelity) than PDP and ALE. The loss monotonically decreases with increasing maximum tree depth for subgroup construction.} 
\label{tab:model-fidelity}
\end{table}

\section{Application}
\label{sec:application}
In the following application, we demonstrate that cs-PDPs and cs-PFI are valuable tools to understand model and data beyond insights given by PFI, PDPs, or ALE plots.
We trained a random forest to predict daily bike rentals \citep{uci} with given weather and seasonal information.
The data ($n=731$, $p=9$) was divided into 70\% training and 30\% test data.

\subsection{Analyzing Feature Dependence}
The features in the bike data are dependent.
For example, the correlation between temperature and humidity is 0.13.
The data contains both categorical and numerical features and we are interested in the multivariate, non-linear dependencies.
Thus, correlation is an inadequate measure of dependence.
We therefore indicate the degree of dependence by showing the extent to which we can predict each feature from all other features in Table \ref{tab:predcor}.
This idea is based on the proportional reduction in loss \citep{cooil1994reliability}.
Per feature, we trained a random forest to predict that feature from all other features.
We measured the proportion of loss explained to quantify the dependence of the respective feature on all other features.
For numerical features, we used the R-squared measure.
For categorical features, we computed $1 - MMCE(y_{class}, rf(X)) / MMCE(y_{class},x_{mode})$, where $MMCE$ is the mean misclassification error, $y_{class}$ the true class, $rf()$ the classification function of the random forest and $x_{mode}$ the most frequent class in the training data.
We divided the training data into two folds and trained the random forest on one half.
Then, we computed the proportion of explained loss on the other half and vice versa.
Finally, we averaged the results.
The feature \enquote{work} can be fully predicted by weekday and holiday.
Season, temperature, humidity and weather can be partially predicted and are therefore not independent.

\begin{table}[ht]
\centering
\begin{tabular}{lllllllll}
  \hline
  \hline
season & yr & holiday & weekday & temp & hum & work & weather & wind \\ 
  45\% & 8\% & 29\% & 14\% & 66\% & 43\% & 100\% & 46\% & 12\% \\ 
   \hline
\end{tabular}
\caption{Percentage of loss explained by predicting a feature from the remaining features with a random forest.} 
\label{tab:predcor}
\end{table}

\subsection{cs-PDPs and cs-PFI}
\begin{figure}
\centering
\includegraphics[width=\columnwidth]{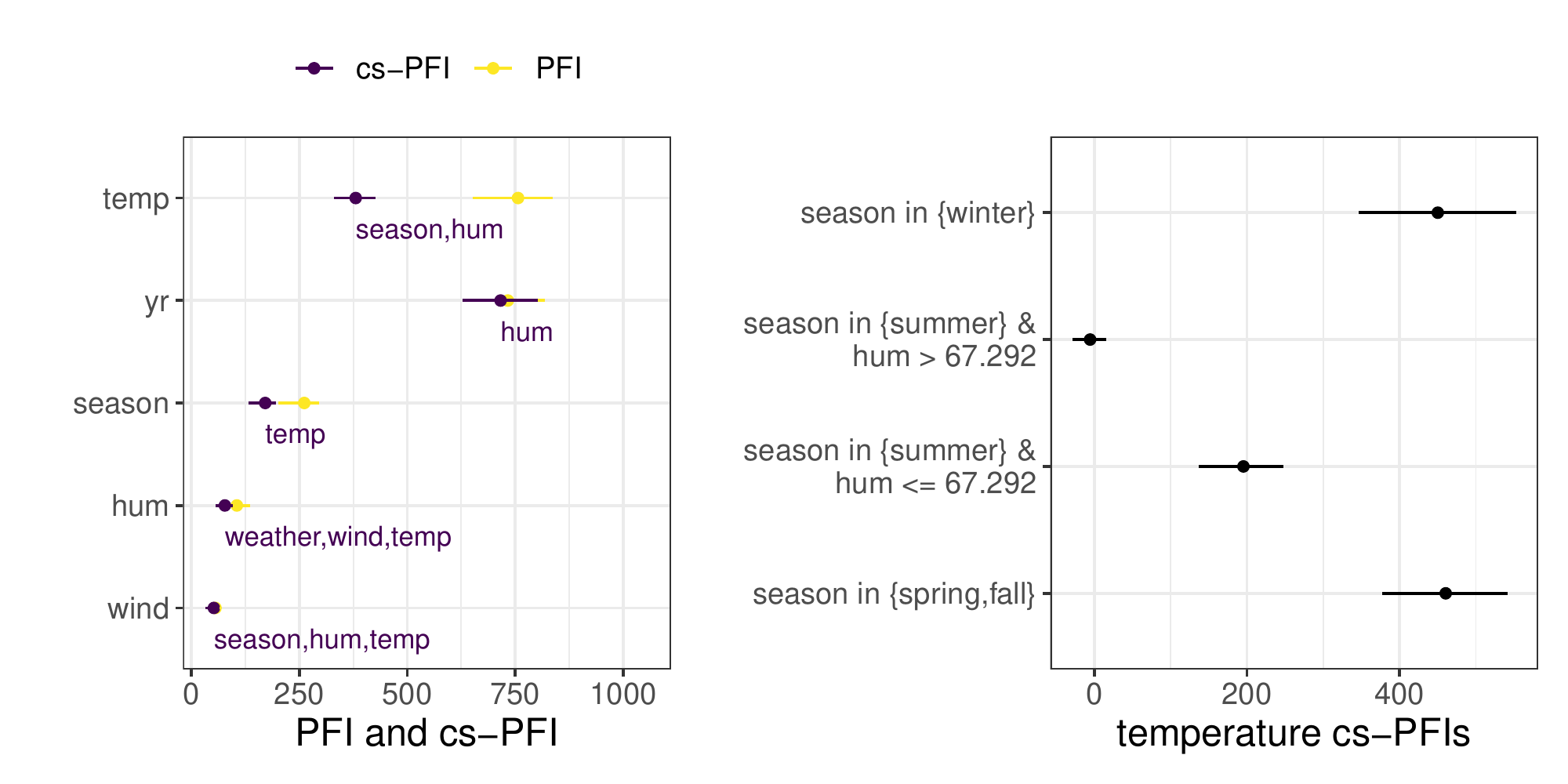} \caption[Left]{\textbf{Left:} Comparison of PFI and cs-PFI for a selection of features. For cs-PFI we also show the features that constitute the subgroups.  \textbf{Right:} Local cs-PFI of temperature within subgroups. The temperature feature is important in spring, fall and winter, but neglectable on summer days, especially humid ones.}\label{fig:compare-fimp-bike}
\end{figure}

To construct the subgroups, we used transformation trees with a maximum tree depth of 2 which  limited the number of possible subgroups to 4.
\begin{figure}
\centering
\includegraphics[width=0.7\columnwidth]{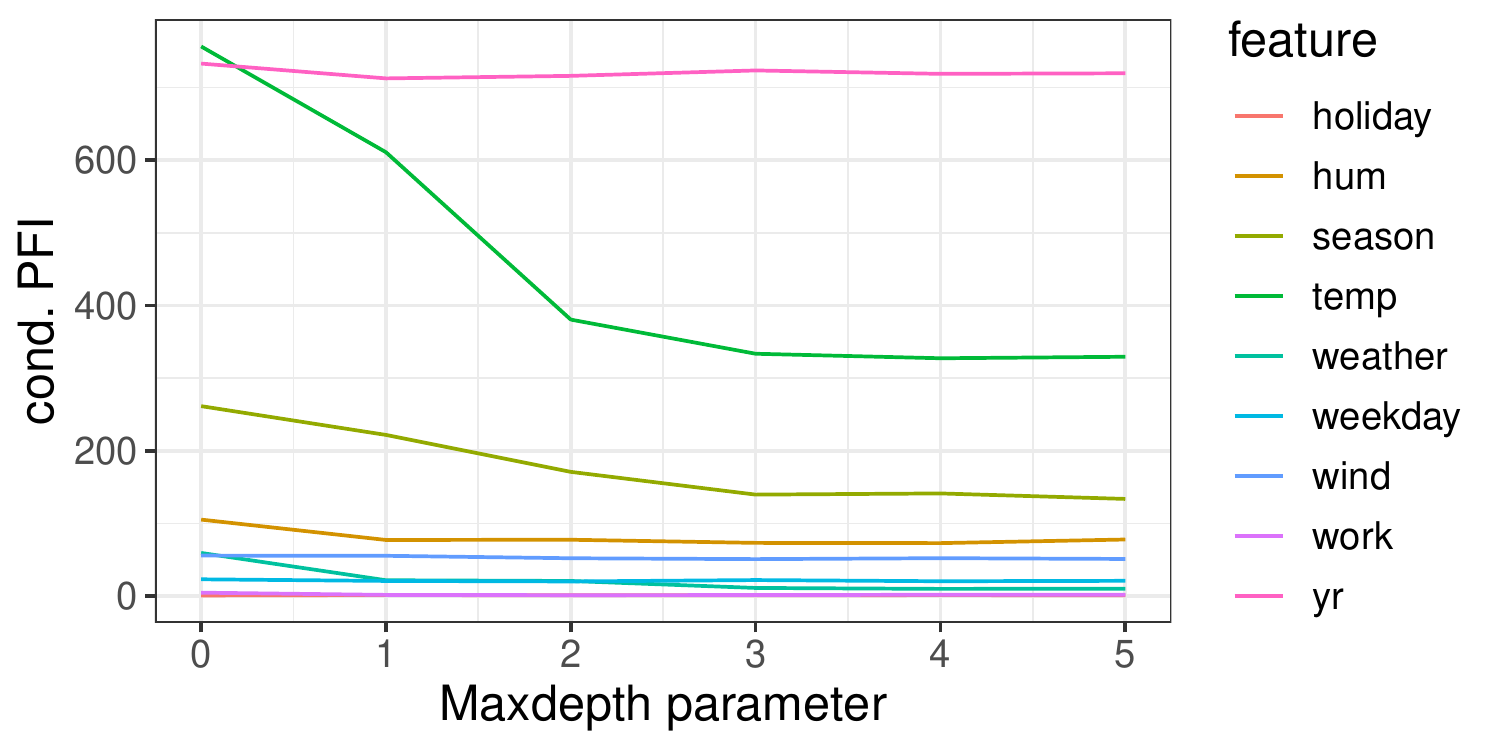}
\caption{Conditional feature importance by increasing  maximum depth of the trees.}
\label{fig:compare-fimp-bike-depth}
\end{figure}
Figure~\ref{fig:compare-fimp-bike-depth} shows that for most features the biggest change in the estimated conditional PFI happens when moving from a maximum depth of 0 (= marginal PFI) to a depth of 2.
This makes a maximum depth of 2 a reasonable trade-off between limiting the number of subgroups and accurately approximating the conditional PFI.
We compared the marginal and conditional PFI for the bike rental predictions, see Figure \ref{fig:compare-fimp-bike}.

The most important features, according to (marginal) PFI, were temperature and year.
For the year feature, the marginal and conditional PFI are the same.
Temperature is less important when we condition on season and humidity.
The season already holds a lot of information about the temperature, so this is not a surprise.
When we know that a day is in summer, it is not as important to know the temperature to make a good prediction.
On humid summer days, the PFI of temperature is zero.
However, in all other cases, it is important to know the temperature to predict how many bikes will be rented on a given day.
The disaggregated cs-PFI in a subgroup can be interpreted as ``How important is the temperature, given we know that the season and the humidity".

\begin{figure}
\includegraphics[width=\columnwidth]{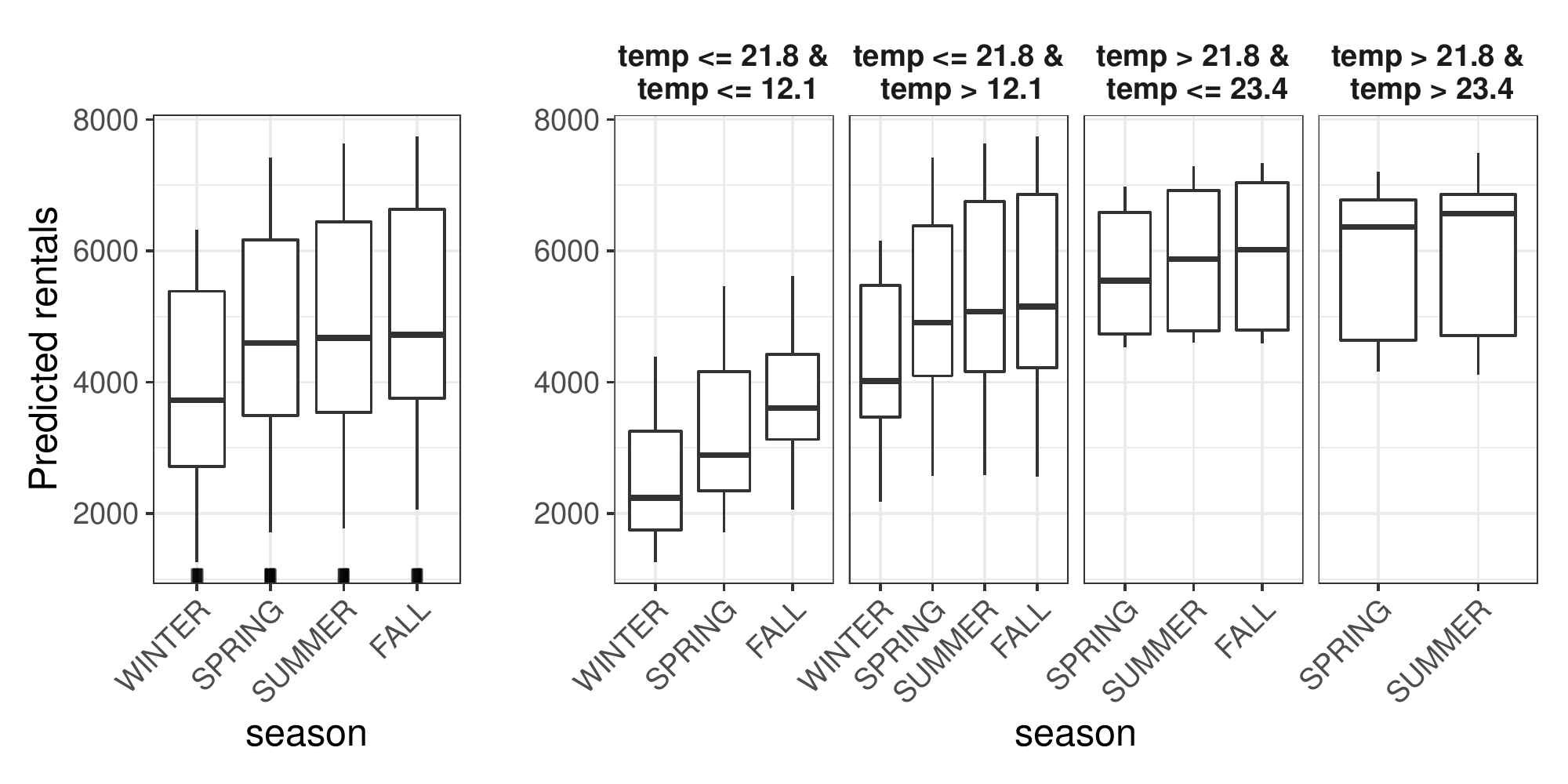} \caption[Effect of season on predicted rentals.]{Effect of season on predicted rentals. \textbf{Left:} PDP. \textbf{Right:} Local cs-PDPs. The cs-PDPs are conditioned on temperature, in which the tree split at 21.5 and at 9.5. }\label{fig:bike-effects-cat}
\end{figure}

Both ALE and PDP show a monotone increase of predicted bike rentals up until a temperature of 25 $^{\circ}$C and a decrease beyond that.
The PDP shows a weaker negative effect of very high temperatures which might be caused by extrapolation: High temperature days are combined with e.g. winter.
A limitation of the ALE plot is that we should only interpret it locally within each interval that was used to construct the ALE plot.
In contrast, our cs-PDP is explicit about the subgroup conditions in which the interpretation of the cs-PDP is valid and shows the distributions in which the feature effect may be interpreted.
The local cs-PDPs in subgroups reveal a more nuanced picture:
For humid summer days, the temperature has no effect on the bike rentals, and the average number of rentals are below that of days with similar temperatures in spring, fall and drier summer days.
The temperature has a slightly negative effect on the predicted number of bike rentals for dry summer days (humidity below 70.75).
The change in intercepts of the local cs-PDP can be interpreted as the effect of the grouping feature (season).
The slope can be interpreted as the temperature effect within a subgroup.\\
We also demonstrate the local cs-PDPs for the season, a categorical feature.
Figure \ref{fig:bike-effects-cat} shows both the PDP and our local cs-PDPs.
The normal PDP shows that on average there is no difference between spring, summer and fall and only slightly less bike rentals in winter.
The PDP with four subgroups conditional on temperature shows that the marginal PDP is misleading.
The PDP indicates that in spring, summer and fall, around $4500$ bikes are rented and in winter around $1000$ fewer.
The cs-PDPs in contrast show that, conditional on temperature, the differences between the seasons are much greater, especially for low temperatures.
Only at high temperatures is the number of rented bikes similar between seasons.

\section{Discussion}
\label{sec:discussion}

We proposed the cs-PFIs and cs-PDPs, wich are variants of PFI and PDP that work when features are dependent.
Both cs-PFIs and cs-PDPs rely on permutations in subgroups based on decision trees.
The approach is simple: Train a decision tree to predict the feature of interest and compute the (marginal) PFI / PDP in each terminal node defined by the decision tree.

Compared to other approaches, cs-PFIs and cs-PDPs enable a human comprehensible grouping, which carries information how dependencies affect feature effects and importance.
As we showed in various experiments, our methods are on par or outperform other methods in many dependence settings.
We therefore recommend using cs-PDPs and cs-PFIs to analyze feature effects and importances when features are dependent.
However, due to their construction with decision trees, cs-PFIs and cs-PDPs do not perform well when the feature of interest depends on many other features, but only if it depends on a few features.
We recommend analyzing the dependence structure beforehand, using the imputation approach with random forests in the case of multiple dependencies, and cs-PFIs in all other cases.

Our framework is flexible regarding the choice of partitioning and we leave the evaluation of the rich selection of possible decision tree and decision rules approaches to future research.

\textbf{Reproducibility:}
All experiments were conducted using \textit{mlr} \citep{mlr3} and R \citep{rlang}.
We used the \textit{iml} package \citep{iml} for ALE and PDP, \textit{party}/\textit{partykit} \citep{hothorn2015partykit} for CVIRF and \textit{knockoff} \citep{knockoff} for Model-X knockoffs.
The code for all experiments is available at \url{https://github.com/christophM/paper_conditional_subgroups}.

\begin{acknowledgements}
This project is funded by the Bavarian State Ministry of Science and the Arts,  by the Bavarian Research Institute for Digital Transformation (bidt) and supported by the German Federal Ministry of Education and Research (BMBF) under Grant No. 01IS18036A and by the German Research Foundation (DFG), Emmy Noether Grant
437611051.
The authors of this work take full responsibilities for its content.
\end{acknowledgements}

\bibliographystyle{spbasic}

\newpage

\vskip 0.2in

\bibliography{Bib}

\newpage

\appendix

\section{Decompose conditional PFI into cs-PFIs}
\label{app:decompose-cpfi}

Assuming a perfect construction of $G_j$, it holds that $X_j \perp X_{-j} | G_j$ and also that $X_j \perp G_j | X_{-j}$ (as $G_j$ is a compression of $X_{-j}$). Therefore
\begin{equation}
  P(X_j|X_{-j}) = P(X_j|X_{-j}, G_j) = P(X_j|G_j).
\end{equation}
When we sample the replacement $\tilde{x}_j^{(i)}$for an $x_j^{(i)}$ from the marginal within a group ($P(X_j|G_j=g_j^{(i)})$, e.g., via permutation) we also sample from the conditional $P(X_j|X_{-j}=x_{-j}^{(i)})$. Every data point from the global sample can therefore equivalently be seen as a sample from the marginal within the group, or as a sample from the global conditional distribution.\\
As follows, the weighted sum of marginal subgroup PFIs coincides with the conditional PFI (cPFI).\\
\begin{align}
  cPFI &= \sum_{i = 1}^{n} \frac{1}{n} \left( L(f (\tilde{x}_j^{(i)}, x_{-j}^{(i)}), y^{(i)}) - L(\fh(x_j^{(i)}, x_{-j}^{(i)}), y^{(i)})\right)\\
  &= \sum_{k = 1}^{K} \frac{n^k}{n} \sum_{i \in \mathcal{G}_k}{}\frac{1}{n^k} \left( L(f (\tilde{x}_j^{(i)}, x_{-j}^{(i)}), y^{(i)}) - L(\fh(x_j^{(i)}, x_{-j}^{(i)}), y^{(i)})\right)\\\\
  &= \sum_{k=1}^{K} \frac{n^k}{n} \textit{PFI}^k
\end{align}

\section{Expectation and Variance of the PFI in a Subgroup}
\label{app:exp-var-cspfi}
We show that under feature independence the PFI and a PFI in an arbitrary subgroup have the same expected value and the subgroup $k$ PFI has a higher variance.
Let $\tilde{L}^{(i)} = \frac{1}{M}\sum_{m=1}^M L(y^{(i)}, \fh(\tilde{x}_j^{m(i)}, x_{-j}^{(i)}) $ and $L^{(i)} = L(y^{(i)}, \fh(x_j^{m(i)}, x_{-j}^{(i)})$.
\begin{proof}
  \begin{align*}
       \E_{X_{-j}}[PFI_j] &= \E_{X_{-j}}\left[\frac{1}{n}\sum_{i=1}^n (\tilde{L}^{(i)} - L^{(i)})\right] \\
       & = \E_{X_{-j}}[\tilde{L}^{(i)} - L^{(i)}]\\
     \E[PFI_j^k]_{X_{-j}} &= \E_{X_{-j}}\left[\frac{1}{n_k}\sum_{i: x^{(i)} \in \mathcal{G}_j^k} (\tilde{L}^{(i)} - L^{(i)})\right]\\
       &= \frac{1}{n_k}\E_{X_{-j}}\left[\sum_{i: x^{(i)} \in \mathcal{G}_j^k} (\tilde{L}^{(i)} - L^{(i)})\right] \\
       &= \frac{1}{n_k} n_k \E_{X_{-j}}\left[(\tilde{L}^{(i)} - L^{(i)})\right] \\
       &=  \E_{X_{-j}}[PFI_j] \\
  \end{align*}
  \begin{align*}
   \mathbb{V}_{X_{-j}}\left[PFI_j\right]
   & = \mathbb{V}_{X_{-j}}\left[\frac{1}{n}\sum_{i=1}^n (\tilde{L}^{(i)} - L^{(i)})\right] \\
   & = \frac{1}{n^2} n \mathbb{V}_{X_{-j}}\left[\tilde{L}^{(i)} - L^{(i)}\right] \\
   & = \frac{1}{n} \mathbb{V}_{X_{-j}}\left[\tilde{L}^{(i)} - L^{(i)})\right] \\
   \mathbb{V}_{X_{-j}}\left[PFI_j^k]\right]
   & = \mathbb{V}_{X_{-j}}\left[\frac{1}{n^k}\sum_{i=1}^{n^k} (\tilde{L}^{(i)} - L^{(i)})\right] \\
   & = \frac{1}{n_k^2} n_k \mathbb{V}_{X_{-j}}\left[\tilde{L}^{(i)} - L^{(i)}\right] \\
   & = \frac{1}{n_k} \mathbb{V}_{X_{-j}}\left[\tilde{L}^{(i)} - L^{(i)})\right] \\
\frac{\mathbb{V}_{X_{-j}}[PFI^k_j]}{\mathbb{V}_{X_{-j}}\left[PFI_j\right]} & = \frac{n}{n_k} \\
   \end{align*}

\end{proof}

\section{Expectation and Variance of the PDP in a Subgroup}
\label{app:exp-var-cspdp}
We show that under feature independence the PDP and a PDP in an arbitrary subgroup have the same expected value and the subgroup $k$ PDP has a higher variance.
\begin{proof}
  \begin{align*}
   \E_{X_{-j}}[PDP_j(x)]  &= \E_{X_{-j}}\left[\fh(x,X_{-j})\right] \\
   \E_{X_{-j}}[PDP^k_j(x)] &= \E_{X_{-j}} \left[\frac{1}{n_k}\sum_{i=1}^{n_k} \fh(x,x_{-j}^{(i)}))\right] =
   \frac{1}{n_k} n_k \E_{X_{-j}}\left[\fh(x,X_{-j})\right] =\\
    & = \E_{X_{-j}}\left[\fh(x,X_{-j})\right]
  \end{align*}
  \begin{align*}
   \mathbb{V}_{X_{-j}}\left[PDP_j(x)\right]
   & = \mathbb{V}_{X_{-j}}\left[\frac{1}{n}\sum_{i=1}^n \fh(x,x_{-j}^{(i)})\right] \\
   & = \frac{1}{n^2} n \mathbb{V}_{X_{-j}}\left[\fh(x,X_{-j})\right] \\
   & = \frac{1}{n} \mathbb{V}_{X_{-j}}\left[\fh(x,X_{-j})\right] \\
   \mathbb{V}_{X_{-j}}\left[PDP^k_j(x)\right]
   & = \mathbb{V}_{X_{-j}}\left[\frac{1}{n_k}\sum_{i=1}^{n_k} \fh(x,x_{-j}^{(i)})\right] \\
   & = \frac{1}{n_k^2} n_{k_j} \mathbb{V}_{X_{-j}}\left[\fh(x,X_{-j})\right] \\
   & = \frac{1}{n_k} \mathbb{V}_{X_{-j}}\left[\fh(x,X_{-j})\right] \\
   \frac{\mathbb{V}_{X_{-j}}[PDP^k_j(x)]}{\mathbb{V}_{X_{-j}}\left[PDP_j(x)\right]} & = \frac{n}{n_k} \\
  \end{align*}

\end{proof}

\newpage
\section{cPFI Ground Truth Scenario II}
\label{app:simulation-rf}
This chapter contains the results for the conditional PFI ground truth simulation, scenario II with an intermediate random forest.

\begin{table}

\caption{\label{tab:mses-ex2}MSE comparing estimated and true conditional PFI (for random forest, scenario II). Legend: impute rf: Imputation with a random forest, ko: Model-X knockoffs, mPFI: (marginal) PFI, tree cart: cs-permutation based on CART, tree trtr: cs-permutation based on transformation trees, CVIRF: conditional variable importance for random forests.}
\centering
\begin{tabular}[t]{lrrrrrr}
\toprule
setting & cs-PFI (cart) & cs-PFI (trtr) & cvirf & impute rf & ko & mPFI\\
\midrule
\addlinespace[0.3em]
\multicolumn{7}{l}{\textbf{independent}}\\
\hspace{1em}n=300, p=10 & 0.26 & 0.28 & 0.22 & 0.27 & 0.25 & 0.27\\
\hspace{1em}n=300, p=90 & 0.19 & 0.17 & 0.14 & 0.18 & 0.19 & 0.17\\
\hspace{1em}n=3000, p=10 & 0.07 & 0.07 & 1.39 & 0.07 & 0.06 & 0.08\\
\hspace{1em}n=3000, p=90 & 0.08 & 0.08 & 1.37 & 0.08 & 0.08 & 0.08\\
\addlinespace[0.3em]
\multicolumn{7}{l}{\textbf{linear}}\\
\hspace{1em}n=300, p=10 & 1.79 & 1.69 & 0.45 & 1.87 & 1.10 & 7.11\\
\hspace{1em}n=300, p=90 & 1.93 & 1.88 & 1.36 & 4.25 & 2.93 & 7.06\\
\hspace{1em}n=3000, p=10 & 0.29 & 0.22 & 5.41 & 0.25 & 0.40 & 6.80\\
\hspace{1em}n=3000, p=90 & 0.32 & 0.24 & 6.98 & 1.66 & 0.26 & 7.02\\
\addlinespace[0.3em]
\multicolumn{7}{l}{\textbf{multi. lin.}}\\
\hspace{1em}n=300, p=10 & 667.79 & 744.48 & 275.58 & 335.40 & 377.35 & 726.15\\
\hspace{1em}n=300, p=90 & 972.42 & 1098.74 & 301.26 & 823.89 & 1473.67 & 1065.26\\
\hspace{1em}n=3000, p=10 & 715.41 & 625.99 & 1790.45 & 114.71 & 454.26 & 1017.53\\
\hspace{1em}n=3000, p=90 & 974.37 & 945.19 & 5090.09 & 532.44 & 110.94 & 1416.30\\
\addlinespace[0.3em]
\multicolumn{7}{l}{\textbf{non-linear}}\\
\hspace{1em}n=300, p=10 & 1.40 & 1.29 & 1.37 & 3.96 & 12.35 & 18.51\\
\hspace{1em}n=300, p=90 & 1.06 & 1.03 & 2.05 & 6.77 & 2.38 & 12.32\\
\hspace{1em}n=3000, p=10 & 0.17 & 0.16 & 6.53 & 1.55 & 15.29 & 17.56\\
\hspace{1em}n=3000, p=90 & 0.15 & 0.14 & 9.09 & 3.28 & 8.00 & 11.30\\
\bottomrule
\end{tabular}
\end{table}

\newpage
\begin{figure}[H]
  \includegraphics[width=\columnwidth]{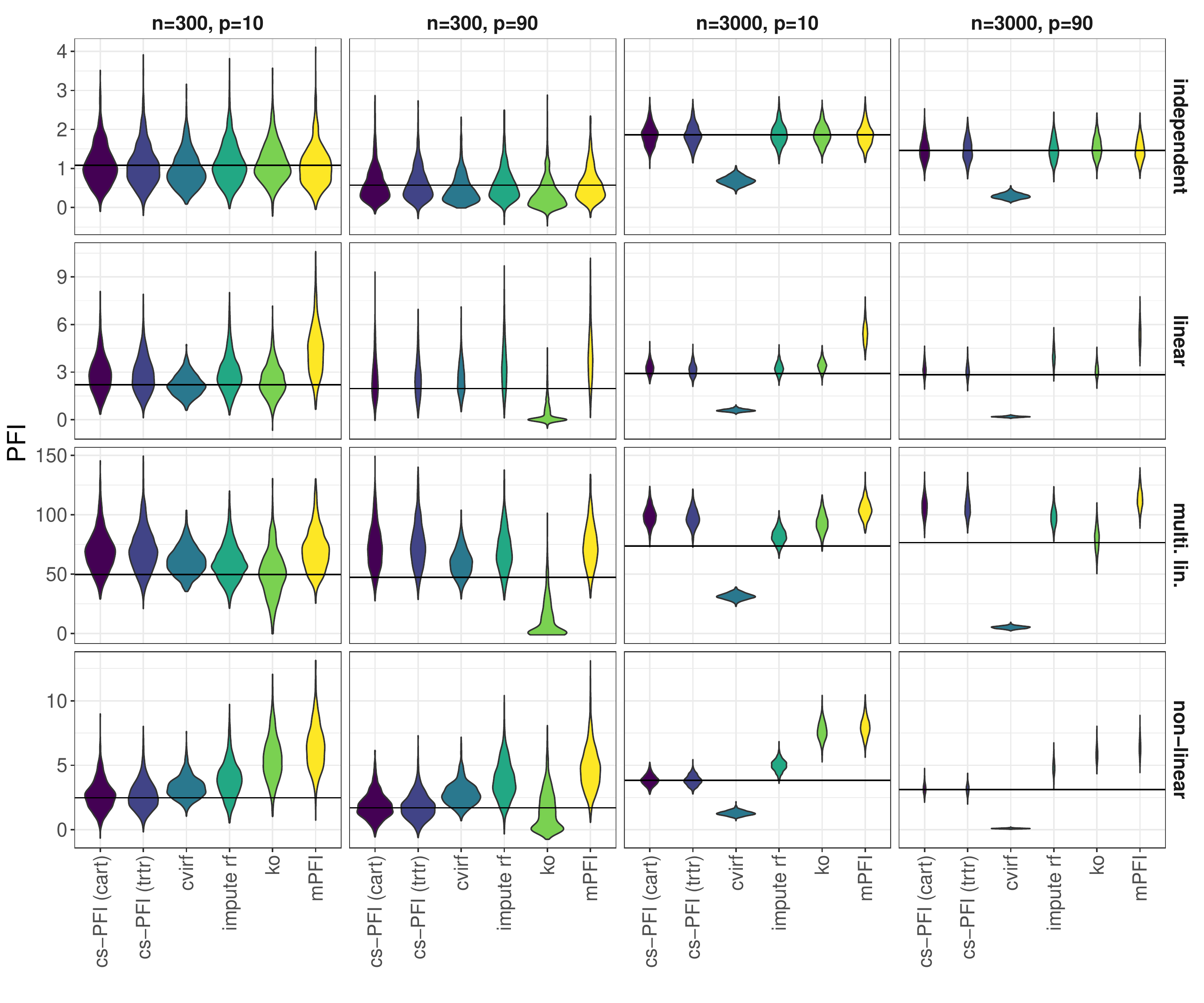} \caption{Experiment (II) comparing various conditional PFI approaches with an intermediary a random forest against the true conditional PFI based on the data generating process.}
  \label{fig:true-importance-all-ex2}
\end{figure}
\newpage

\section{Data Fidelity on OpenML-CC18 data sets}
\label{app:data-fidelity}

An overview of data sets from the OpenML-CC18 benchmarking suit. We used a subset of 42 out of 72 data sets with 7 to 500 continuous features.

\begin{table}[ht]
\centering
\begingroup\small
\begin{tabular}{rlrrr}
  \hline
OpenML ID & Name & No. Obs. & No. numerical feat. & No. feat. \\ 
  \hline
1049 & pc4 & 1458 &  38 &  38 \\ 
  1050 & pc3 & 1563 &  38 &  38 \\ 
  1053 & jm1 & 10880 &  22 &  22 \\ 
  1063 & kc2 & 522 &  22 &  22 \\ 
  1067 & kc1 & 2109 &  22 &  22 \\ 
  1068 & pc1 & 1109 &  22 &  22 \\ 
   12 & mfeat-factors & 2000 & 217 & 217 \\ 
   14 & mfeat-fourier & 2000 &  77 &  77 \\ 
  1461 & bank-marketing & 45211 &   8 &  17 \\ 
  1475 & first-order-theorem-proving & 6118 &  52 &  52 \\ 
  1480 & ilpd & 583 &  10 &  11 \\ 
  1486 & nomao & 34465 &  90 & 119 \\ 
  1487 & ozone-level-8hr & 2534 &  73 &  73 \\ 
  1494 & qsar-biodeg & 1055 &  42 &  42 \\ 
  1497 & wall-robot-navigation & 5456 &  25 &  25 \\ 
   15 & breast-w & 683 &  10 &  10 \\ 
  1501 & semeion & 1593 & 257 & 257 \\ 
  151 & electricity & 45312 &   8 &   9 \\ 
  1510 & wdbc & 569 &  31 &  31 \\ 
   16 & mfeat-karhunen & 2000 &  65 &  65 \\ 
  182 & satimage & 6430 &  37 &  37 \\ 
  188 & eucalyptus & 641 &  15 &  20 \\ 
   22 & mfeat-zernike & 2000 &  48 &  48 \\ 
  23517 & numerai28.6 & 96320 &  22 &  22 \\ 
   28 & optdigits & 5620 &  63 &  65 \\ 
  307 & vowel & 990 &  11 &  13 \\ 
   31 & credit-g & 1000 &   8 &  21 \\ 
   32 & pendigits & 10992 &  17 &  17 \\ 
   37 & diabetes & 768 &   9 &   9 \\ 
  40499 & texture & 5500 &  41 &  41 \\ 
  40701 & churn & 5000 &  17 &  21 \\ 
  40966 & MiceProtein & 552 &  78 &  82 \\ 
  40979 & mfeat-pixel & 2000 & 241 & 241 \\ 
  40982 & steel-plates-fault & 1941 &  28 &  28 \\ 
  40984 & segment & 2310 &  19 &  20 \\ 
  40994 & climate-model-simulation-crashes & 540 &  21 &  21 \\ 
   44 & spambase & 4601 &  58 &  58 \\ 
  4538 & GesturePhaseSegmentationProcessed & 9873 &  33 &  33 \\ 
  458 & analcatdata\_authorship & 841 &  71 &  71 \\ 
   54 & vehicle & 846 &  19 &  19 \\ 
    6 & letter & 20000 &  17 &  17 \\ 
  6332 & cylinder-bands & 378 &  19 &  40 \\ 
   \hline
\end{tabular}
\endgroup
\caption{Overview of OpenML CC18 data sets used for the data fidelity experiment.} 
\label{tab:cc18}
\end{table}

\pagebreak
\subsection{Data Fidelity Results}

\begin{figure}[H]
  \includegraphics[width=\columnwidth]{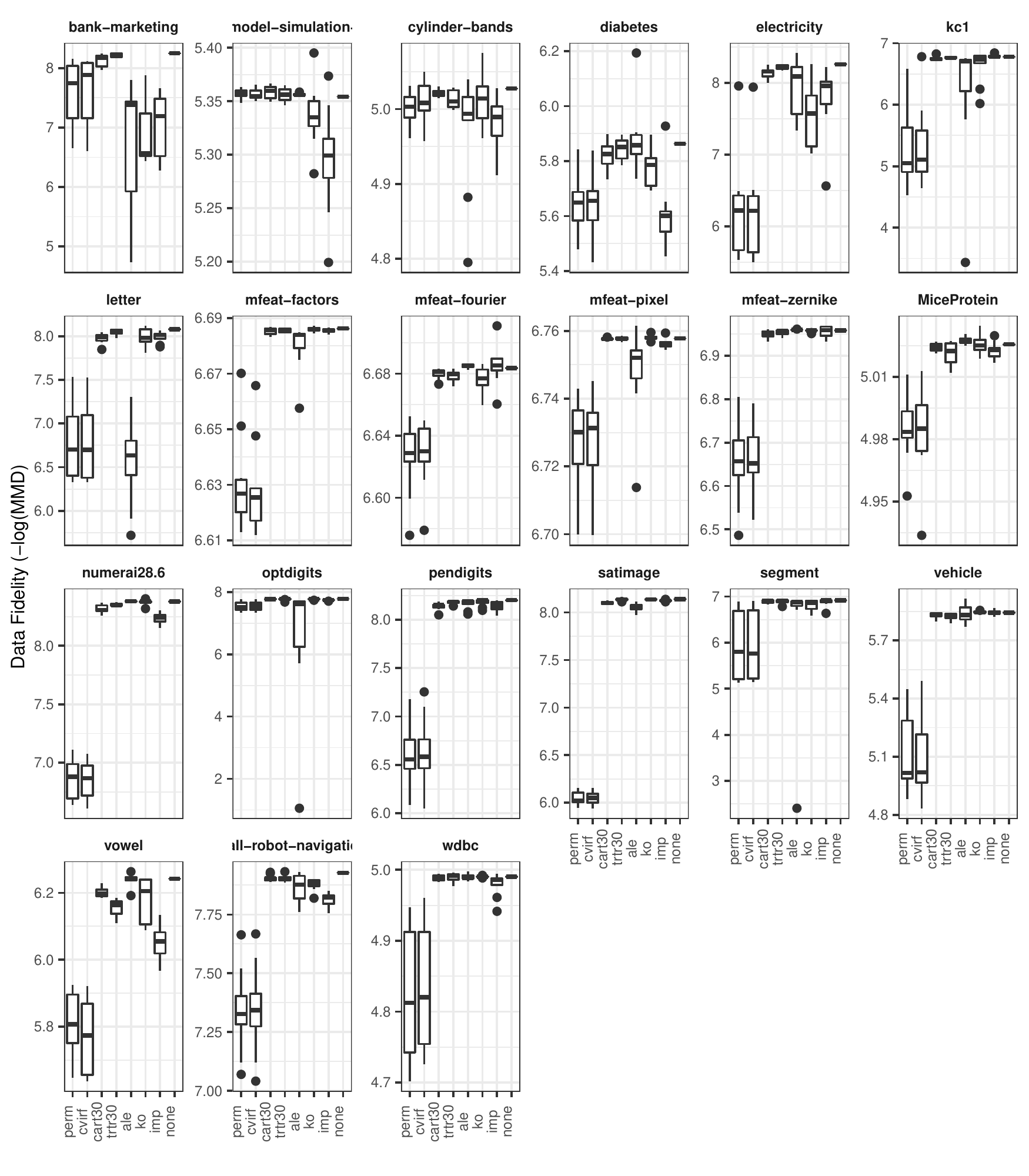} \caption{Data Fidelity experiment with OpenML-CC18 data sets (1/2). Different sampling types are compared: unconditional permutation (perm), cs-permutation (maximal tree depth) with CART (cart30) or transformation trees (trtr30), Model-X knockoffs (ko), data imputation with a random forest (imp), ALE (ale), conditional variable importance for random forests (cvirf) and no permutation (none). Each data point in the boxplot represents one feature and one data set. Results from repeated experiments have been averaged (mean) before using them in the boxplots.}
  \label{fig:data-mmd-all1}
\end{figure}

\begin{figure}[H]
  \includegraphics[width=\columnwidth]{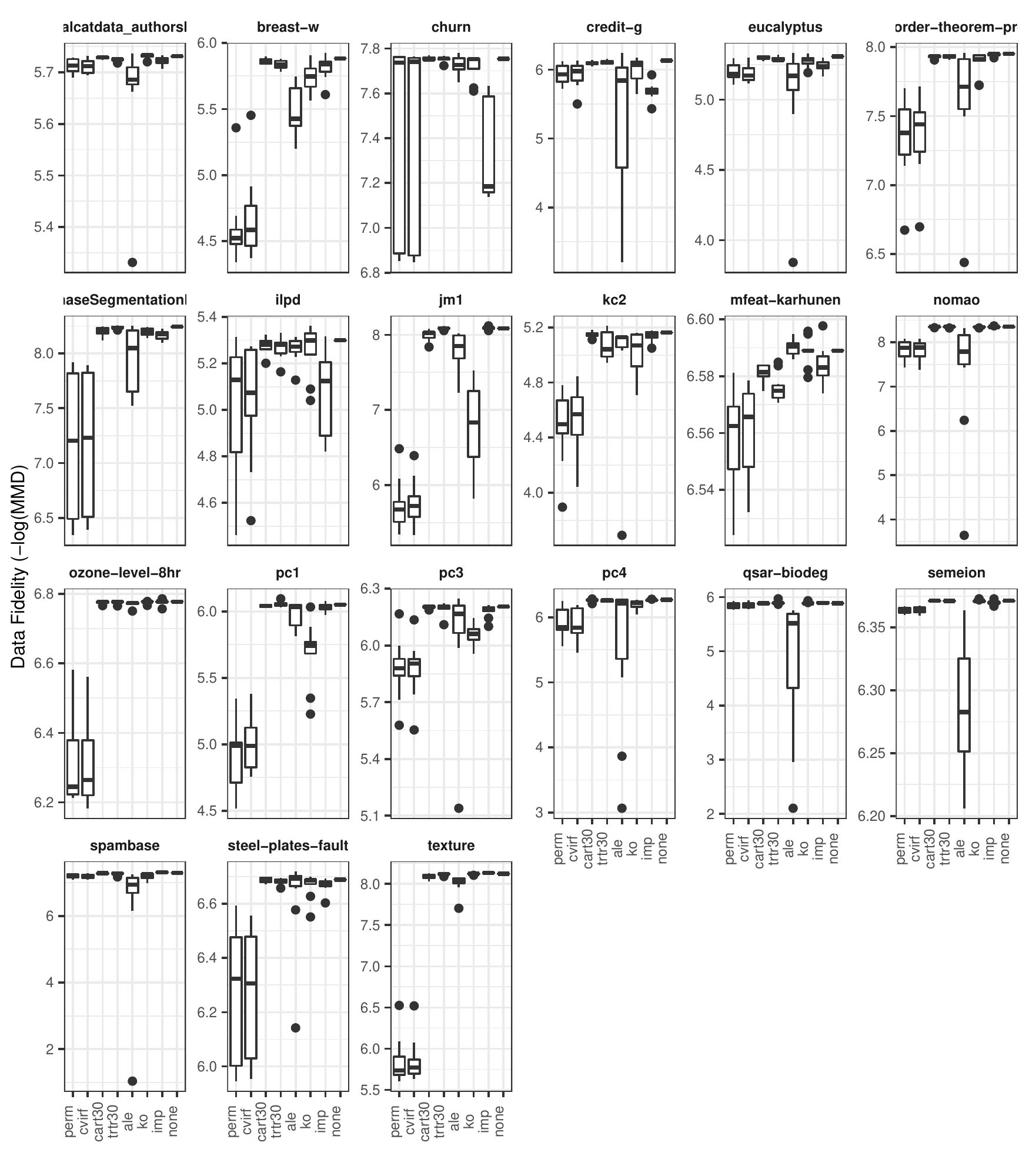} \caption{Data Fidelity experiment with OpenML-CC18 data sets (2/2). Different sampling types are compared: unconditional permutation (perm), cs-permutation (maximal tree depth) with CART (cart30) or transformation trees (trtr30), Model-X knockoffs (ko), data imputation with a random forest (imp), ALE (ale), conditional variable importance for random forests (cvirf) and no permutation (none). Each data point in the boxplot represents one feature and one data set. Results from repeated experiments have been averaged (mean) before using them in the boxplots.}
  \label{fig:data-mmd-all2}
\end{figure}

\begin{figure}[H]
  \includegraphics[width=\columnwidth]{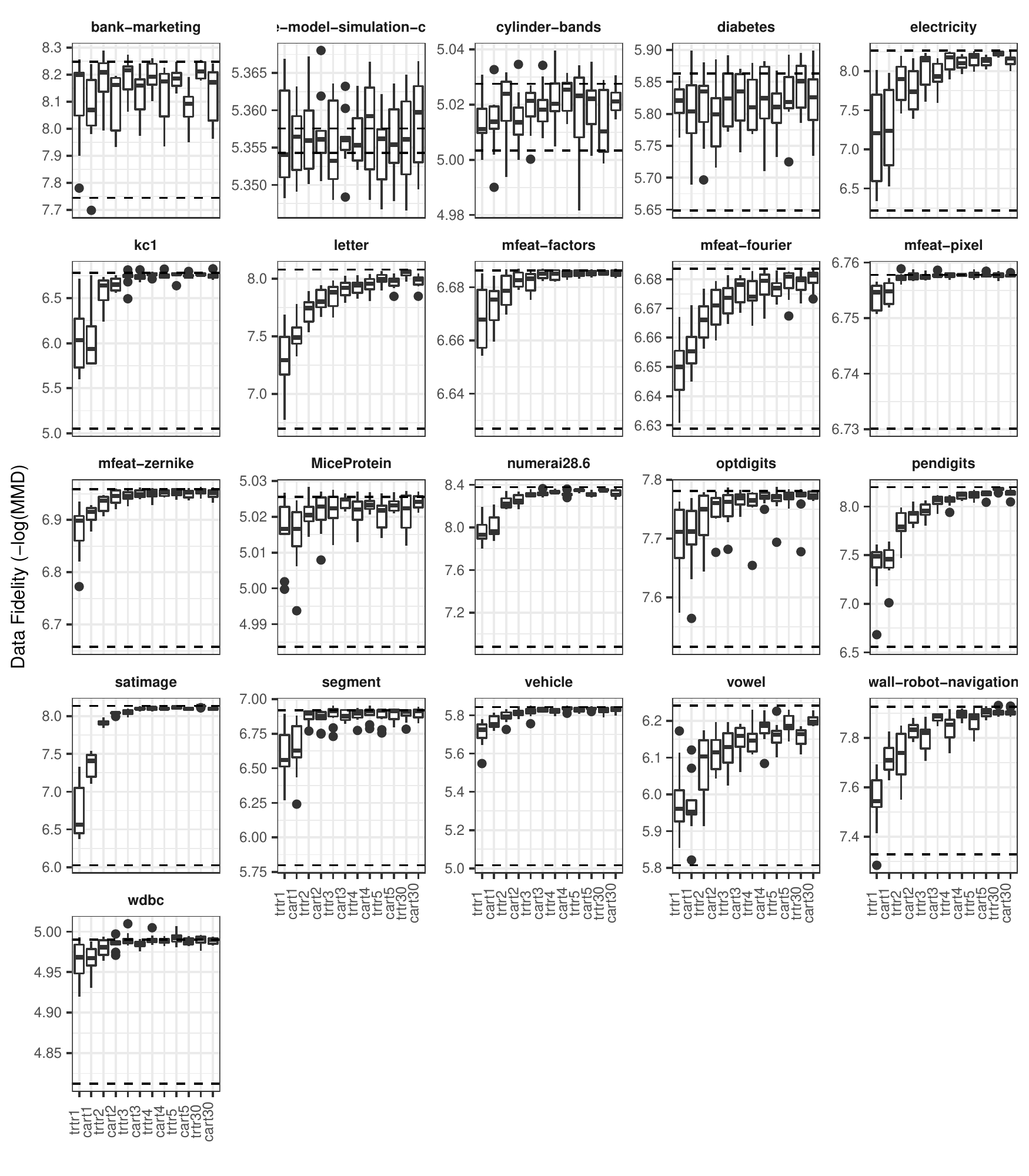} \caption{Data Fidelity experiment with OpenML-CC18 data sets (1/2). Different tree depths and tree types (CART and Transformation Trees) are compared. Unconditional permutation and lack of permutation serve as lower and upper bound for data fidelity and their median data fidelity is plotted as dotted lines. Each data point in the boxplot represents one feature and one data set. Results from repeated experiments have been averaged (mean) before using them in the boxplots.}
\label{fig:data-mmd-tree1}
\end{figure}

\begin{figure}[H]
  \includegraphics[width=\columnwidth]{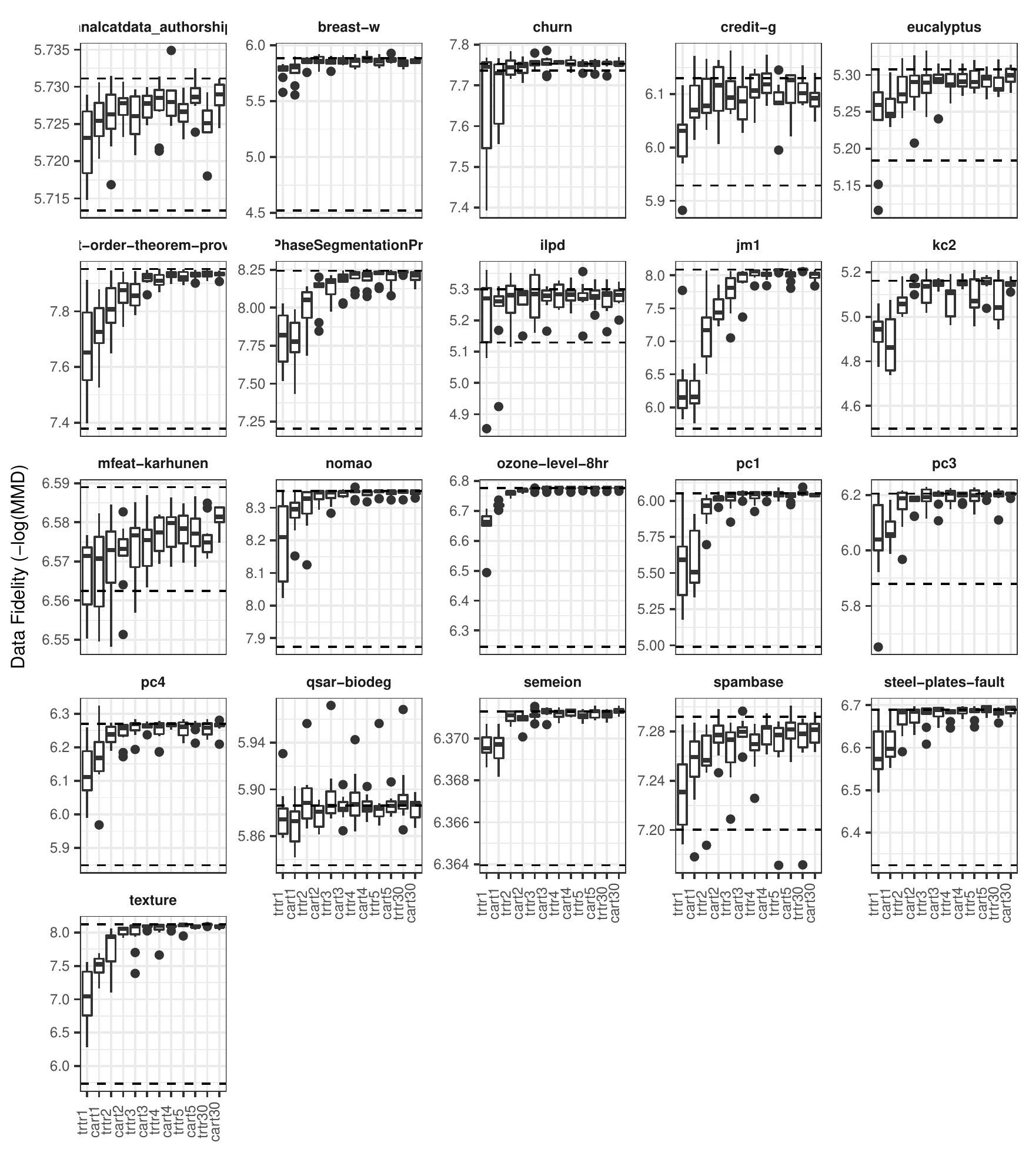} \caption{Data Fidelity experiment with OpenML-CC18 data sets (2/2). Different tree depths and tree types (CART and Transformation Trees) are compared. Unconditional permutation and lack of permutation serve as lower and upper bound for data fidelity and their median data fidelity is plotted as dotted lines. Each data point in the boxplot represents one feature and one data set. Results from repeated experiments have been averaged (mean) before using them in the boxplots.}
\label{fig:data-mmd-tree2}
\end{figure}

\section{Model Fidelity Plots}
\label{sec:model-fid-extra}

\begin{figure}[H]
\includegraphics[width=\textwidth]{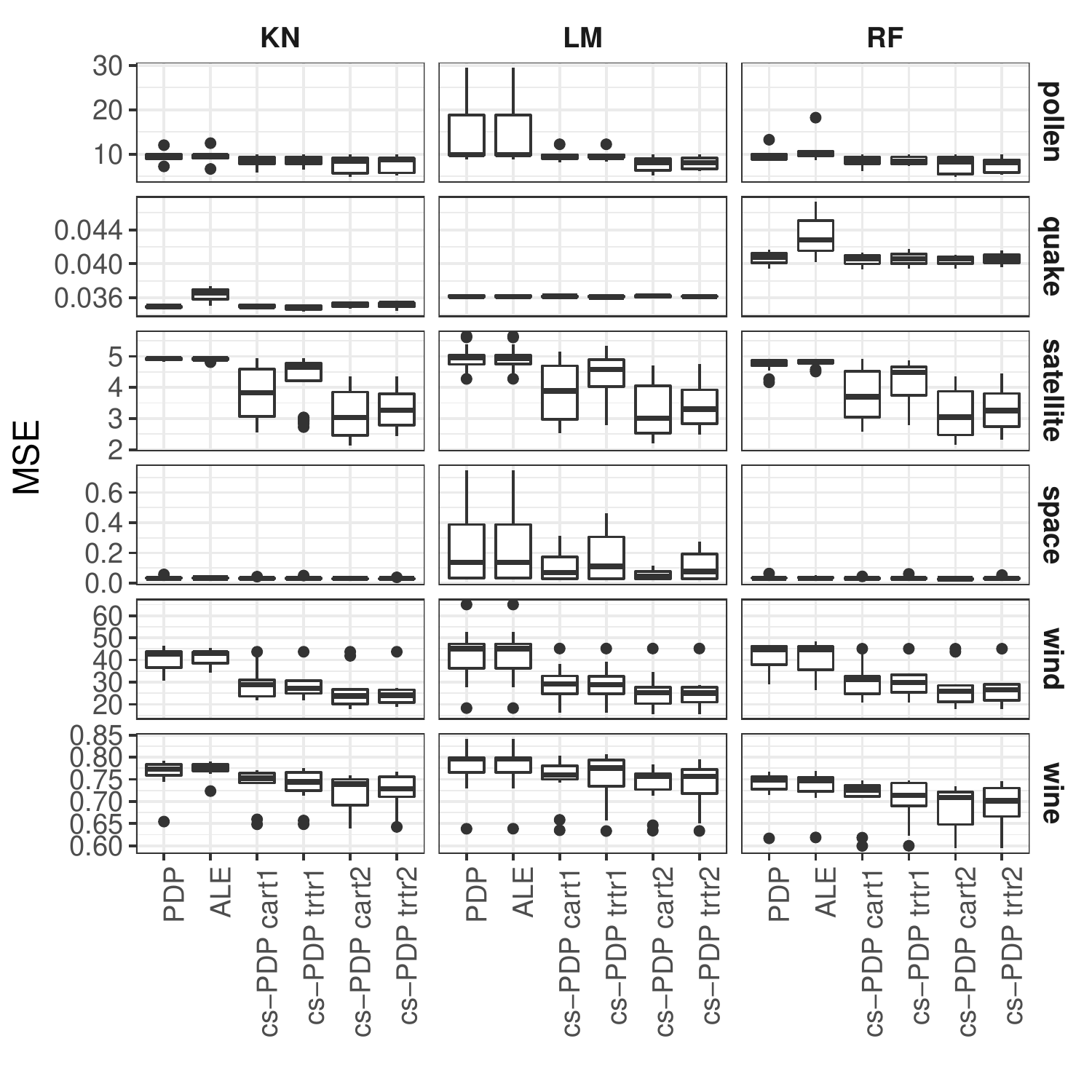} \caption[Comparing the loss between model f and ]{Comparing the loss between model f and various feature effect methods. Each instance in the boxplot is MSE for one feature, summed over the test data.}\label{fig:predictiveness}
\end{figure}

\end{document}